%% file: main.tex
\documentclass[10pt,twocolumn,letterpaper]{article}

\usepackage{cvpr}              % To produce the CAMERA-READY version
\usepackage{algorithm}
\usepackage{booktabs}
\usepackage{multirow}
\usepackage{graphicx}
\usepackage{threeparttable}
\usepackage{algorithmic}
\usepackage{longtable}
\usepackage{array}
\usepackage{caption}
\usepackage{xcolor}
\usepackage{makecell}

\usepackage[table]{xcolor}
\definecolor{rowgray}{RGB}{220,220,220}
\input{preamble}
\definecolor{cvprblue}{rgb}{0.21,0.49,0.74}
\usepackage[pagebackref,breaklinks,colorlinks,allcolors=cvprblue]{hyperref}

\def\paperID{20} % *** Enter the Paper ID here
\def\confName{CVPR}
\def\confYear{2026}
\newcommand{\methodname}{\textbf{DAMP}}

\title{Class Unlearning via Depth-Aware Removal of Forget-Specific Directions}

\author{Arman Hatami\\
Johns Hopkins University\\
{\tt\small ahatami2@jh.edu}
\and
Romina Aalishah\\
Johns Hopkins University\\
{\tt\small raalish1@jh.edu}
\and
Ilya E. Monosov\\
Johns Hopkins University\\
{\tt\small imonoso1@jh.edu}
}

\begin{document}
\maketitle
\input{sec/0_abstract}    
\input{sec/1_intro}
\input{sec/2_relatedworks}
\input{sec/3_Method}
\input{sec/4_experimants}

\input{sec/5_result}

\input{sec/6_limit}
\input{sec/7_conclusion}
{
    \small
    \bibliographystyle{ieeenat_fullname}
    \bibliography{main}
}
%
%\input{sec/X_suppl}
%WARNING: do not forget to delete the %supplementary pages from your submission 
\input{sec/X_suppl}

\end{document}

%% file: sec/0_abstract.tex
\begin{abstract}
Machine unlearning aims to remove targeted knowledge from a trained model without the cost of retraining from scratch. In class unlearning, however, reducing accuracy on forget classes does not necessarily imply true forgetting: forgotten information can remain encoded in internal representations, and apparent forgetting may arise from classifier-head suppression rather than representational removal. We show that existing class-unlearning methods often exhibit weak or negative selectivity, preserve forget-class structure in deep representations, or rely heavily on final-layer bias shifts. We then introduce \methodname{} (Depth-Aware Modulation by Projection), a one-shot, closed-form weight-surgery method that removes forget-specific directions from a pretrained network without gradient-based optimization. At each stage, \methodname{} computes class prototypes in the input space of the next learnable operator, extracts forget directions as residuals relative to retain-class prototypes, and applies a projection-based update to reduce downstream sensitivity to those directions. To preserve utility, \methodname{} uses a parameter-free depth-aware scaling rule derived from probe separability, applying smaller edits in early layers and larger edits in deeper layers. The method naturally extends to multi-class forgetting through low-rank subspace removal. Across MNIST, CIFAR-10, CIFAR-100, and Tiny ImageNet, and across convolutional and transformer architectures, \methodname{} more closely resembles the retraining gold standard than some of the prior methods, improving selective forgetting while better preserving retain-class performance and reducing residual forget-class structure in deep layers.
\end{abstract}

%% file: sec/1_intro.tex
\vspace{-0.5em}
\begin{flushright}
\begin{minipage}{\linewidth}
\itshape
``In the practical use of our intellect, forgetting is as important a function as recollecting.''\\[0.4em]
\raggedleft
--- William James, \textit{The Principles of Psychology}, Vol.~I
\end{minipage}
\end{flushright}
\vspace{-10pt}
\section{Introduction}
\label{sec:intro}

Machine learning models have grown increasingly large and complex, with modern systems often pretrained for extended periods on vast and diverse datasets. After deployment, however, these models may need to be modified without full retraining. In this work we focus on class unlearning: removing the
% a model’s 
knowledge of one or more target classes from a model while preserving performance on the remaining classes ~\cite{cao2015towards,chundawat2023zero,zhou2025decoupled,wang2024machine}.

\begin{figure}[t]
    \centering
    \includegraphics[width=0.85\linewidth]{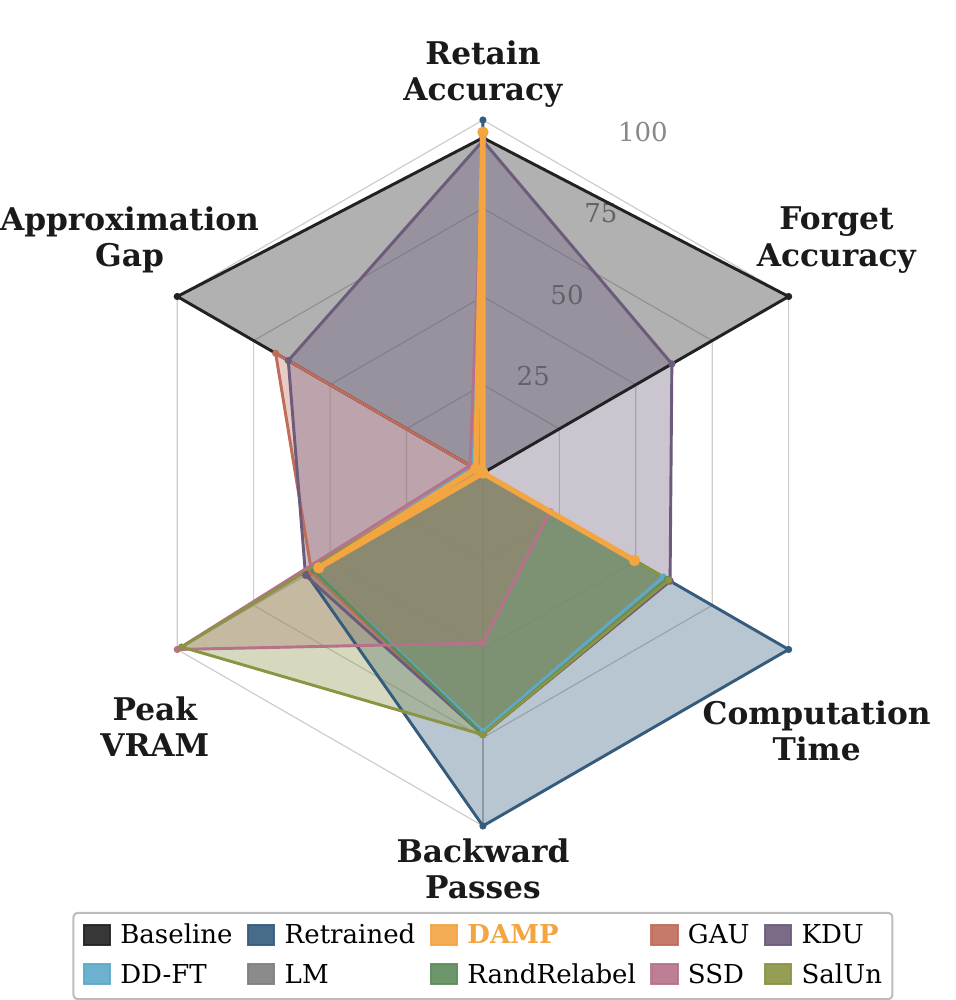}
    \vspace{-8pt}
    \caption{Comparison of \methodname{} performance with the baseline, retrained model, Gradient Ascent Unlearning (GAU), knowledge-distillation unlearning (KDU), Data Deletion Fine-Tuning (DD-FT), Logit Masking~(LM), Random Relabeling~(RandRelabel), Selective Synaptic Dampening~(SSD), and Saliency Unlearning~(SalUn) for a 5-layer CNN~(CNN-5) on CIFAR-10; forget classes 3 (Cat) and 5 (Dog).}
    \label{fig:spider}
    \vspace{-15pt}
\end{figure}

Retraining large models from scratch on datasets that exclude the unwanted forget classes is often impractical due to the computational cost and time required  ~\cite{bourtoule2021machine,wang2024machine,cadet2025deep}. As a result, recent work has developed post-hoc class-unlearning methods that modify trained models directly  ~\cite{chundawat2023zero,kurmanji2023towards,foster2024fast,zhou2025decoupled}.
The objective of these approaches is to make the model behave as if the forget classes had never been learned, while retaining as much useful knowledge as possible about the remaining classes  ~\cite{bourtoule2021machine,wang2024machine}.

Conventionally, most existing work evaluates class unlearning using a tradeoff: reducing performance on the forget classes while preserving performance on the retain classes   ~\cite{bourtoule2021machine,wang2024machine,cadet2025deep}.
Fig.~\ref{fig:spider} summarizes this tradeoff, comparing our method, Depth-Aware Modulation by Projection~(\methodname{}), with retraining and prior unlearning baselines in terms of retain accuracy, forget accuracy, and computational efficiency.
Although informative, the approach of measuring the tradeoff as a metric of unlearning does not fully capture the selectivity of the forgetting process.  This figure illustrates that a given method may lower accuracy on forget classes while also degrading retain-class performance, or it may also preserve overall utility while leaving substantial evidence for forget classes in the model ~\cite{cadet2025deep,hayes2025inexact}.

\begin{figure}[t]
    \centering
    \includegraphics[width=0.85\linewidth]{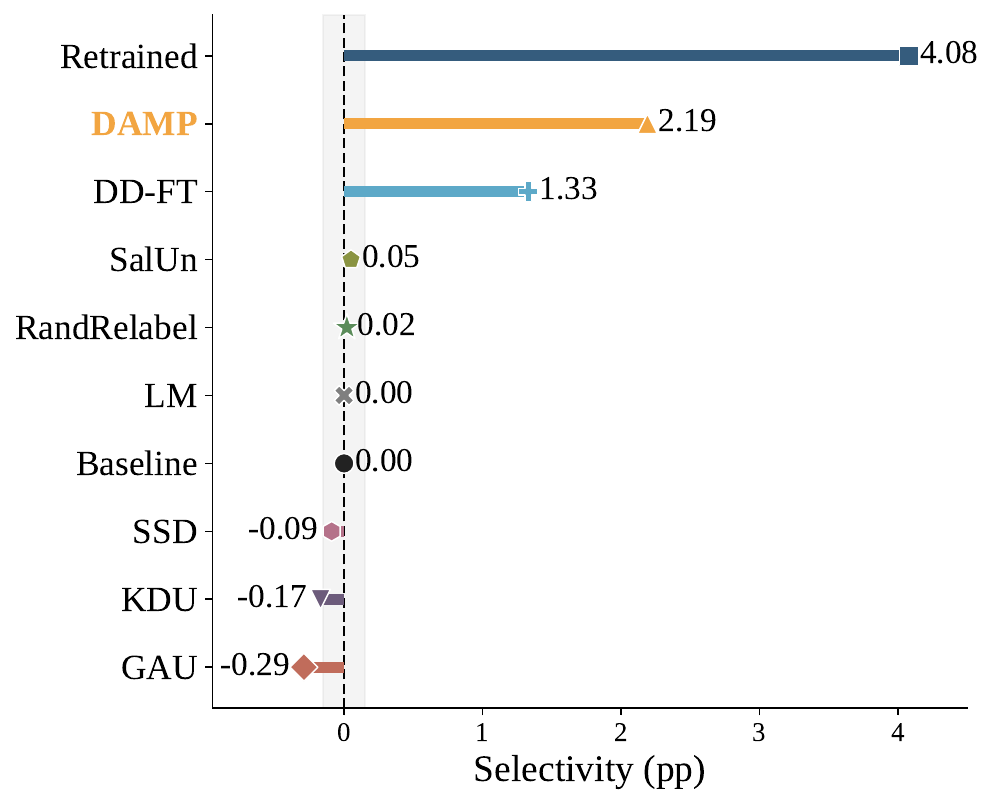}
    \vspace{-10pt}
    \caption{Comparison of \methodname{} Selectivity with the baseline retrained model, Gradient Ascent Unlearning (GAU), knowledge-distillation unlearning (KDU), Data Deletion Fine-Tuning (DD-FT), Logit Masking (LM), Random Relabeling~(RandRelabel), Selective Synaptic Dampening~(SSD), and Saliency Unlearning~(SalUn) for a 5-layer CNN~(CNN-5) on CIFAR-10; forget classes 3 (Cat) and 5 (Dog). Existing methods exhibit weak, near-zero, or negative selectivity, indicating either incomplete forgetting or collateral damage to retained representations. In contrast, \methodname{} substantially improves selectivity and appears closer to the retrained network (a gold standard comparison). The full equation of Selectivity is given in Sec.~\ref{sec:supp_selectivity}.}
    \label{fig:selectivity_intro}
    \vspace{-15pt}
\end{figure}

To further shed light on this distinction and begin to develop a novel method for class forgetting, we first evaluated the selectivity of the forgetting process, measured in percentage points (pp), a probe-based metric that captures how effectively a method suppresses forget-class evidence without degrading retain-class representations. As illustrated in Fig.~\ref{fig:selectivity_intro}, existing methods often exhibit relatively weak, near-zero, or even negative selectivity.
These patterns of weak selectivity indicate forgetting that arises from broadly damaging representations or from suppressing forget classes only at the classifier head. In contrast, high selectivity indicates that forget-class evidence is removed more strongly than retain-class structure is degraded.

\begin{figure*}[t]
    \centering
    \includegraphics[width=0.97\textwidth]{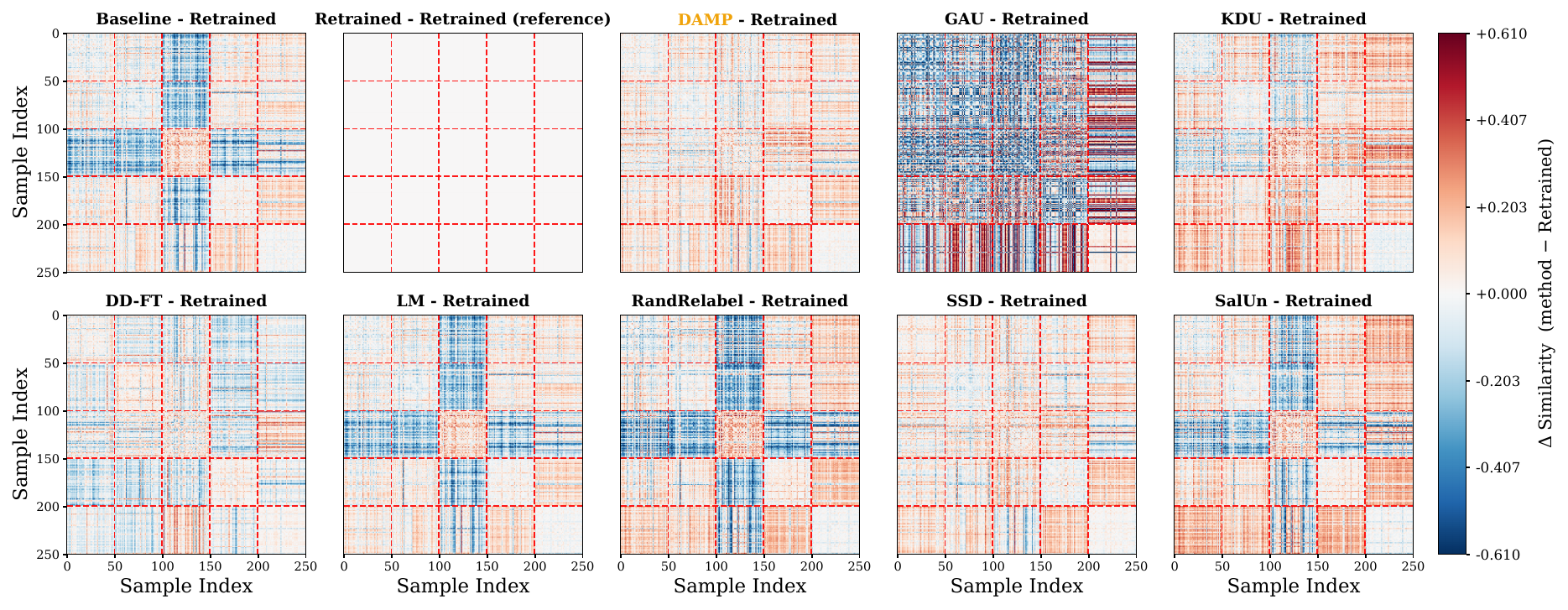}
    \vspace{-10pt}
    \caption{The first panel (top-left) shows the difference between the RDM of the original model that has not been subjected to unlearning (Baseline) and the retrained (gold-standard) model that has been retrained without the \textit{forget} class. The other panels show the difference between the RDM of models subjected to different unlearning methods and the RDM of the retrained model. These unlearning methods are \ \methodname{}, Gradient Ascent Unlearning (GAU), knowledge-distillation unlearning (KDU), Data Deletion Fine-Tuning (DD-FT), Logit Masking (LM), Random Relabeling~(RandRelabel), Selective Synaptic Dampening~(SSD), and Saliency Unlearning~(SalUn) for a 5-layer CNN~(CNN-5) on CIFAR-10; retain classes 4 (Deer), 5 (Dog), 7 (Horse), 8 (Ship) and forget class 6 (Frog). Existing unlearning methods exhibit larger deviations from retraining, either by preserving stronger forget-class structure or by distorting the geometry of retain classes, whereas \methodname{} more closely matches retrained network properties. Original RDM plot can be seen in Fig.~\ref{fig:rdm_org}.}
    \label{fig:rdm_intro}
    \vspace{-15pt}
\end{figure*}

% Weak selectivity is only part of the problem. 
Moreover, even when output-level forgetting appears successful, forgotten information can remain encoded in internal representations~\cite{hayes2025inexact,cadet2025deep}. This issue becomes more pronounced as a function of network depth. Since early layers often capture low-level features shared across classes, whereas deeper layers encode increasingly abstract and class-specific structure ~\cite{yosinski2014transferable}, some residual decodability in shallow layers may not be surprising. 
However, if forget-class structure remains detectable in deeper layers following unlearning, then high-level semantic evidence for forget classes has not truly been removed. This is the case even if the \textit{forgotten} classes are no longer predicted by the output of the network.
As we show in Fig.~\ref{fig:rdm_intro} and Fig.~\ref{fig:tsne}, several existing methods continue to preserve forget-class structure in deep representations of networks relative to what is found in a retrained network, motivating the development of an unlearning procedure that acts beyond the classifier head.

Furthermore, apparent forgetting can arise from changes at the classifier head rather than from genuine removal of forget-class evidence. In particular, shifting the final-layer bias can suppress the prediction of a forget class while leaving its internal representation largely intact. This makes output-level forgetting alone sometimes an unreliable indicator of successful class unlearning ~\cite{cadet2025deep,hayes2025inexact}. We revisit this mechanism in Sec.~\ref{sec:related_work}, and later discuss how several existing methods rely primarily on classifier-head suppression rather than representational removal.

% Based on these observations
To address this issue, we 
% propose
show that class unlearning should modify the parts of the network where class-specific evidence is encoded while preserving shared low-level structure. This motivates an approach that (i) operates across network depth and (ii) applies conservative edits in earlier stages and stronger edits in deeper stages~\cite{yosinski2014transferable}.
% We 
Therefore, we introduce \methodname{} (\underline{D}epth-\underline{A}ware \underline{M}odulation by \underline{P}rojection), a one-shot, closed-form weight-surgery method for class unlearning that removes forget-specific directions from a pretrained network without gradient-based optimization. At each stage, we define an edit space as the input space of the next learnable operator and compute class prototypes in that space. We then extract forget directions as residual components that are not explained by retain-class prototypes, and apply a projection-based update to the next-operator weights to reduce sensitivity to those directions. To limit utility loss, we scale the update at each stage using a parameter-free coefficient derived from forget-versus-retain probe separability together with a deterministic depth ramp, resulting in smaller edits in earlier layers and larger edits in deeper layers. For multi-class forgetting, we compute one direction per forget class, orthonormalize the resulting set, and project out their span, yielding a low-rank extension without introducing additional hyperparameters. Unlike gradient-based unlearning methods, which iteratively optimize a forgetting objective, \methodname{} removes forget-specific directions from internal feature spaces. By targeting deep class-specific structure rather than relying only on output suppression, \methodname{} reduces the risk that forget-class information remains encoded in internal representations, where it may distort the geometry of retained-class features, harm generalization on unseen data, and remain vulnerable to re-expression under alternative heads, downstream fine-tuning, or representation-level analysis. 

Our contributions are as follows: 
\begin{itemize}
    \item We show that current class-unlearning methods often fail to achieve selective forgetting: they either damage retained knowledge, preserve forget-class structure across depth, or rely on output-layer suppression.
    \item We propose \methodname{}, an architecture-agnostic, closed-form projection-surgery method for class unlearning that removes forget-specific directions in the input space of the next learnable operator.
    \item We introduce a parameter-free, depth-aware scaling rule based on probe separability and extend the method to multi-class forgetting via low-rank subspace removal.
\end{itemize}

%% file: sec/2_relatedworks.tex
\section{Related Work}
\label{sec:related_work}

Machine unlearning spans a broad range of settings, including sample-level deletion, concept removal, and class-level forgetting ~\cite{cao2015towards,bourtoule2021machine,wang2024machine,cadet2025deep,baumhauer2022machine}. In this paper, we focus on class-unlearning, where the goal is to remove knowledge of one or more target classes while preserving performance on the remaining classes. Within this setting, existing methods can be broadly grouped into a few families. Retraining-based methods remove the forget classes and retrain from scratch, providing the idealized 
% ``as-if-never-seen'' reference 
baseline but at prohibitive computational cost ~\cite{Kim2022CVPR,bourtoule2021machine,chourasia2023forget}. Gradient-ascent-based methods increase the loss on the forget classes to push predictions away from the forgotten labels; representative examples include zero-shot and unified gradient-based formulations ~\cite{chundawat2023zero,huang2024unified,mavrothalassitis2025ascent}. These methods are simple and can be effective, but they may be unstable and often do not explicitly target the internal representations that encode forget-class evidence. Optimization-based methods combine forgetting and retention objectives, often through distillation, regularization, or auxiliary remain-data constraints ~\cite{kurmanji2023towards,zhou2025decoupled,bonato2024retain}. Such formulations offer flexibility, but they operate through objective tradeoffs rather than directly specifying what class-specific structure should be removed. Relabeling-based methods replace forget labels with retain-class labels and then fine-tune the model, thereby altering the output mapping without necessarily removing class-specific evidence from the representation space ~\cite{graves2021amnesiac,he2025towards,li2023random}. Representation-based methods act more directly on hidden features through pruning, saliency-guided edits, synaptic dampening, or other internal modifications, to localize forgetting more precisely~\cite{jia2023model,fan2023salun,foster2024fast,sepahvand2025selective,kodge2024deep}.

Despite these differences, existing class-unlearning methods often exhibit three recurring limitations. First, forgetting can come at the cost of degraded retain-class performance, a tradeoff that appears across many approximate unlearning methods and benchmark settings ~\cite{kurmanji2023towards,cadet2025deep,hayes2025inexact}. As 
in Fig.~\ref{fig:selectivity_intro} many methods in our experiments can exhibit weak, near-zero, or even negative selectivity, indicating that forgetting is not cleanly separated from collateral damage to retained representations. Second, forget-class information can remain recoverable from internal representations even when top-level forgetting metrics improve, echoing broader evidence that deep networks can preserve discriminative features across layers and transfer settings ~\cite{yosinski2014transferable,zhu2024decoupling,sepahvand2025selective}. As shown in Fig.~\ref{fig:rdm_intro} and further illustrated by the t-SNE~\cite{maaten2008visualizing} visualization in Fig.~\ref{fig:tsne}, representational dissimilarity matrices for prior methods often retain substantial block structure associated with forget classes, suggesting that forget-class evidence is still encoded even when output predictions are reduced. Third, some methods can, in some cases, produce apparent forgetting primarily by changing the decision boundary or classifier head, rather than clear removal of internal class evidence~\cite{chen2023boundary,he2025towards,zhu2024decoupling}.

This distinction matters because output-level suppression and representational removal are not the same phenomenon. A model may stop predicting a forget class while still retaining features that make that class distinguishable in its hidden representations. In that case, the model has reduced output access to the forget class without fully removing the underlying internal evidence. For class unlearning, this can make output-level metrics alone an incomplete measure of success ~\cite{hayes2025inexact,zhu2024decoupling,sepahvand2025selective}.

\begin{figure}[t]
    \centering
    \includegraphics[width=0.9\linewidth]{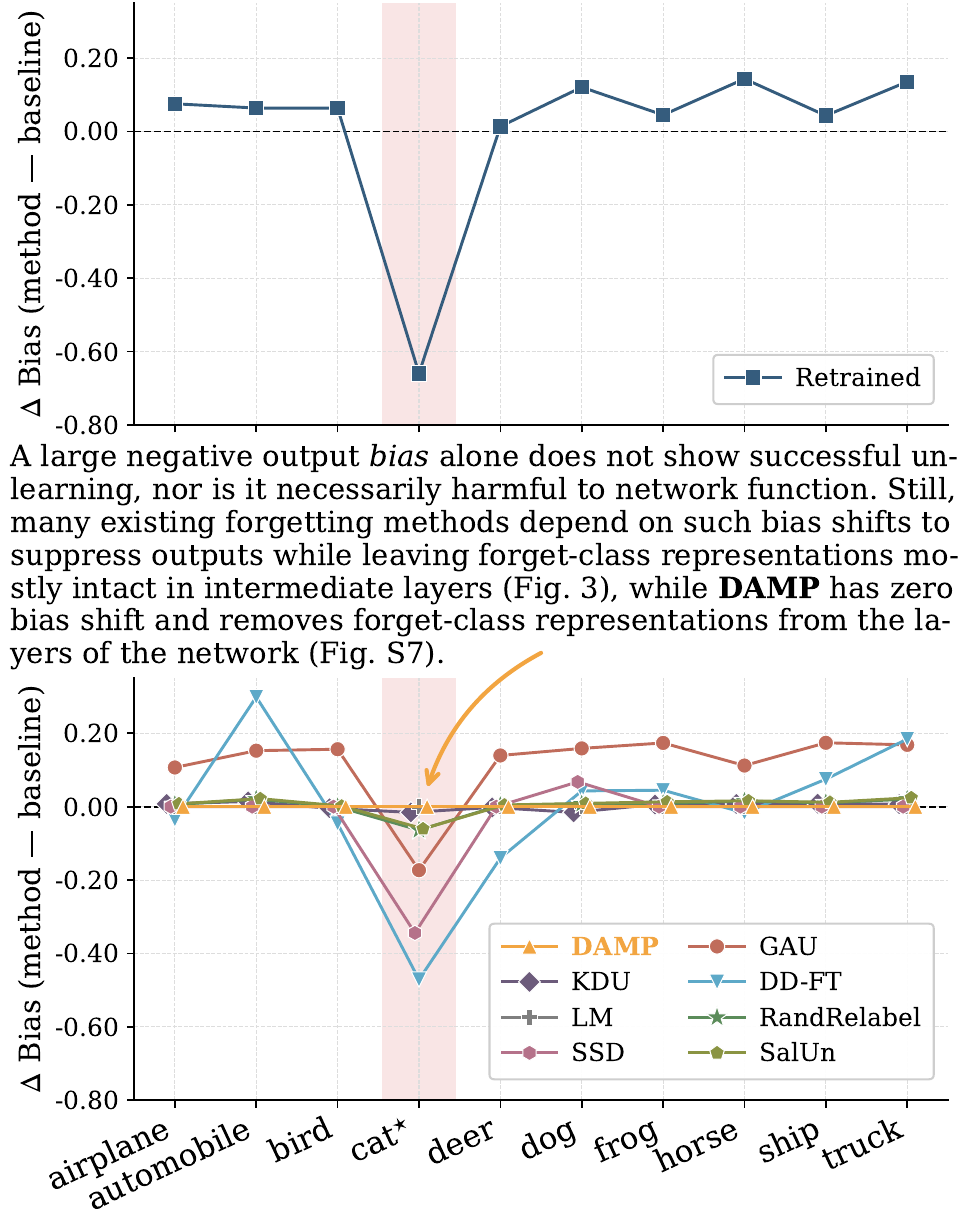}
    \vspace{-10pt}
    \caption{Comparison of \methodname{} bias shift with the baseline, retrained model, Gradient Ascent Unlearning (GAU), knowledge-distillation unlearning (KDU), Data Deletion Fine-Tuning (DD-FT), Logit Masking (LM), Random Relabeling~(RandRelabel), Selective Synaptic Dampening~(SSD), and Saliency Unlearning~(SalUn) for a 5-layer CNN~(CNN-5) on CIFAR-10; forget class 3 (Cat). The asterisk marks the forget class. Existing unlearning methods often drive the forget-class bias strongly negative, reducing predictions through classifier-head suppression rather than genuine removal of internal evidence. In contrast, \methodname{} requires less reliance on bias manipulation. More analysis on bias shift is provided in~\ref{sec:supp_bias}.}
    \label{fig:bias_related}
    \vspace{-20pt}
\end{figure}

One indicator of this behavior is the bias term in the final classification layer. In a linear classifier,
% $\mathbf{w}^\top \mathbf{x} + b$
the bias term contributes a class-dependent offset to the pre-softmax score. Reducing the bias for a forget class makes that class less likely to be predicted, even if the corresponding feature representations remain largely intact~\cite{hinton2015distilling}. As shown in Fig.~\ref{fig:bias_related}, several existing methods in our evaluation drive the forget-class bias towards negative values, suggesting that part of their \textit{apparent }forgetting in fact arises from classifier-head suppression rather than substantial representational removal.

Our method is most closely related to representation-level and post-hoc parameter-editing approaches ~\cite{fan2023salun,foster2024fast,jia2023model,sepahvand2025selective,kodge2024deep}, but differs from them in two key ways. First, rather than editing parameters through iterative optimization, pruning, or neuron-level intervention, we identify at each stage a forget-specific direction; or, in the multi-class case, a low-rank forget subspace; relative to the span of retain-class prototypes. Second, we edit the next learnable operator by right-projecting its weight matrix to reduce sensitivity to those directions, which weakens the ability of downstream computation to exploit forget-class evidence. Because the procedure is one-shot, closed-form, and depth-aware, it directly targets internal class-specific structure while aiming to preserve shared low-level features and retain-class utility~\cite{yosinski2014transferable}.

%% file: sec/3_Method.tex
\section{Method}
\label{sec:method}

\begin{figure*}[t]
    \centering
    \includegraphics[width=0.7\textwidth]{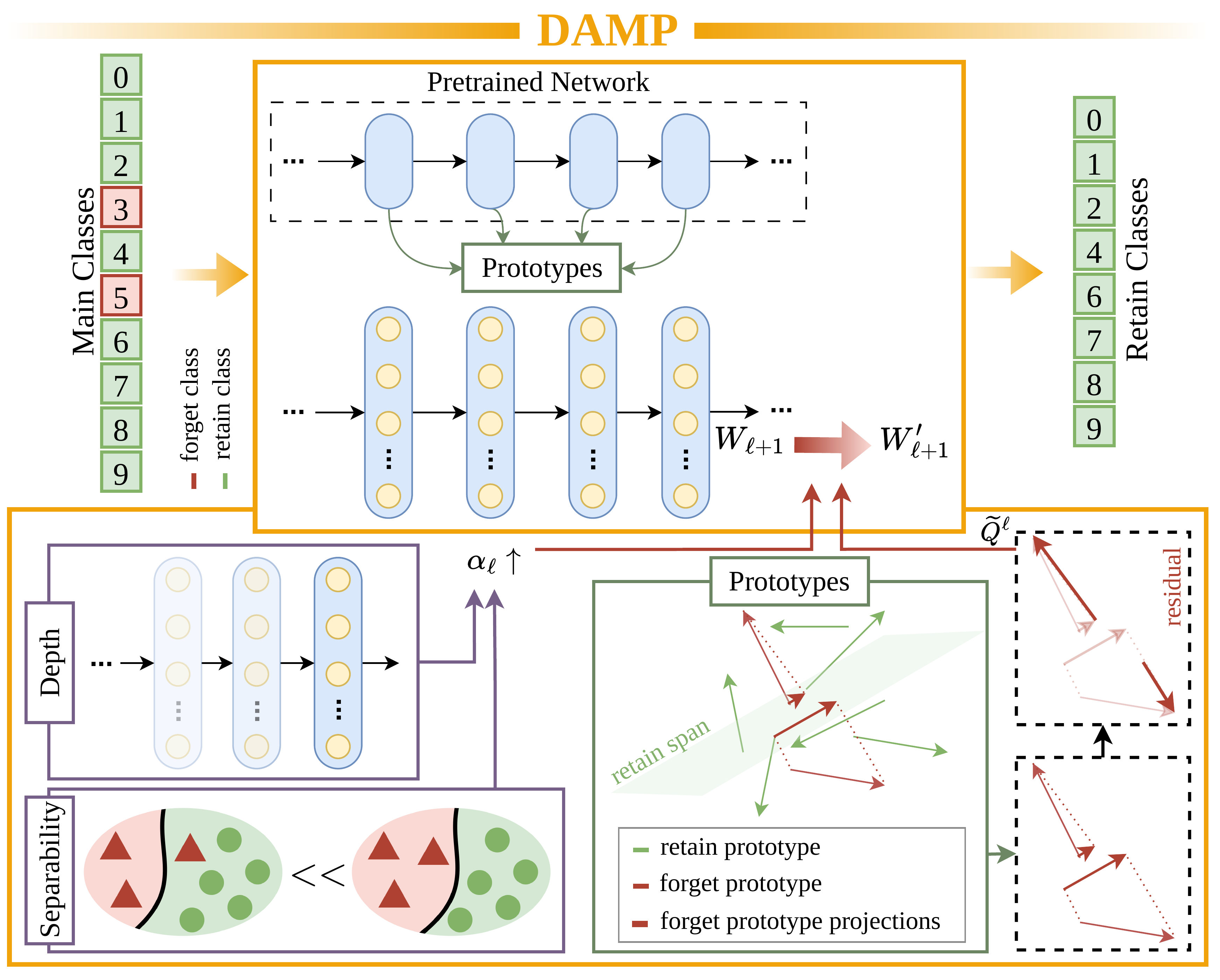}
    \vspace{-10pt}
    \caption{Overview of the proposed \methodname{}. Starting from a pretrained network, we compute class prototypes in the edit space of each stage. For each forget class, its prototype is decomposed into a component explained by the retain span and a forget residual. The resulting residual directions are orthonormalized. In parallel, a layer coefficient $\alpha_\ell$ is computed from depth and forget-retain separability. These two quantities define the projection update applied to the weights of the next layer.}
    \label{fig:method_overview}
    \vspace{-15pt}
\end{figure*}

\paragraph{Setting.}
Let $F(\cdot;\theta)$ be a pretrained classifier.
We are given a set of forget classes $\mathcal{F}$ and a retain set $\mathcal{D}_r$ containing classes $\mathcal{C}_r$ with $\mathcal{C}_r\cap\mathcal{F}=\emptyset$, following the standard retain-forget formulation used in class-unlearning methods~\cite{kodge2024deep}.
For each forget class $f\in\mathcal{F}$, let $\mathcal{D}_f$ denote its examples.
We produce edited parameters $\theta'$ via a one-shot, closed-form weight update. For the probe we use the union-of-forgets set $\mathcal{D}_{\mathcal{F}}=\bigcup_{f\in\mathcal{F}}\mathcal{D}_f$.
All required statistics, including class prototypes and probe accuracies, are computed once from the pretrained model; the subsequent weight edit uses no gradients and does not require iterative re-estimation, in contrast to iterative forgetting procedures~\cite{kodge2024deep}.
    
Fig.~\ref{fig:method_overview} illustrates the full procedure. Starting from a pretrained network, we compute class prototypes in the edit space of each stage. Each forget prototype is then projected onto the span of the retain prototypes, so that the retained component is separated from the forget residual. The residual directions are stacked and orthonormalized to define the forget subspace $\widetilde{Q}^\ell$ at that stage. In parallel, a scalar coefficient $\alpha_\ell$ is computed so that deeper stages and stages with stronger forget-retain separability receive stronger edits. These two quantities are then combined in a closed-form projection update that reduces the sensitivity of the next operator to the forget subspace while preserving the remaining representational space.

\vspace{-10pt}

\paragraph{Edited stages and next-operator view.}
Consider a sequence of feature-producing stages indexed by $\ell\in\{1,\dots,L\}$.
For each stage $\ell$, let $T_{\ell+1}$ denote the next operator that consumes the representation produced at stage $\ell$ and has a learnable weight tensor.
We define an edit-space vector $z^\ell(x)\in\mathbb{R}^{d_\ell}$ as the actual input to $T_{\ell+1}$, after reshaping if needed so that $T_{\ell+1}$ is linear in $z^\ell$:
\begin{equation}
T_{\ell+1}(z^\ell) \;=\; W^{\ell+1}z^\ell + b^{\ell+1}.
\end{equation}
This definition is architecture-agnostic: for any operator with a weight tensor, we flatten its input dimension(s) to obtain $z^\ell$ and the corresponding matrix $W^{\ell+1}\in\mathbb{R}^{m_\ell\times d_\ell}$.
This next-operator view is similar in spirit to direct model-editing methods that modify selected linear mappings in pretrained networks via explicit parameter updates~\cite{meng2022locating,meng2022mass}.

\paragraph{Class prototypes in edit space.}
For each retain class $c\in\mathcal{C}_r$ and each stage $\ell$, we compute the edit-space prototype
\begin{equation}
\boldsymbol{\mu}_c^\ell \;=\; \frac{1}{|\mathcal{D}_c|}\sum_{x\in\mathcal{D}_c}z^\ell(x).
\label{eq:proto_retain}
\end{equation}
For each forget class $f\in\mathcal{F}$ we compute a separate prototype
\begin{equation}
\boldsymbol{\mu}_f^\ell \;=\; \frac{1}{|\mathcal{D}_f|}\sum_{x\in\mathcal{D}_f}z^\ell(x).
\label{eq:proto_forget_multi}
\end{equation}
Using class means as prototype representations is standard in feature-space classification and metric-based learning~\cite{snell2017prototypical}.
All prototypes are computed using the pretrained model (before any edits) and are kept fixed during surgery.
Recomputing prototypes after each layer update would require an iterative procedure with repeated forward passes and would change the target directions across steps. We note that our method can be easily extended beyond the utilization of the standard class means as prototype representations to capture more complex class distributions if application requires. 

\vspace{-10pt}

\paragraph{Forget directions via retain-span residual (multi-1D).}
For each stage $\ell$, stack retain prototypes as columns:
\begin{equation}
R^\ell \;=\; \big[\boldsymbol{\mu}_{c_1}^\ell,\dots,\boldsymbol{\mu}_{c_{|\mathcal{C}_r|}}^\ell\big]\in\mathbb{R}^{d_\ell\times|\mathcal{C}_r|}.
\end{equation}
For each forget class $f\in\mathcal{F}$, we compute a retain-span residual
\begin{equation}
d_f^\ell
\;=\;
\boldsymbol{\mu}_f^\ell
-
R^\ell(R^\ell)^{\dagger}\boldsymbol{\mu}_f^\ell,
\label{eq:residual_multi}
\end{equation}
and normalize
\begin{equation}
q_f^\ell \;=\; \frac{d_f^\ell}{\|d_f^\ell\|_2},
\label{eq:q_multi}
\end{equation}
skipping any $f$ for which $\|d_f^\ell\|_2$ is below a small numerical tolerance.
Our construction is motivated by subspace-based editing and orthogonal representation editing methods, which isolate target information in a low-dimensional direction or subspace while reducing interference with unrelated behavior~\cite{cai2024edit,uppaal2025modeleditingrobustdenoised,biswas2025cure}.

Let $Q^\ell$ stack the resulting directions as columns:
\begin{equation}
Q^\ell \;=\; \big[q_{f_1}^\ell,\dots,q_{f_K}^\ell\big]\in\mathbb{R}^{d_\ell\times K},
\end{equation}
where $K\le |\mathcal{F}|$ after skipping near-zero residuals.
We then form an orthonormal basis of $\mathrm{span}(Q^\ell)$ using QR decomposition:
\begin{equation}
\widetilde{Q}^\ell \;=\; \mathrm{orth}(Q^\ell),
\label{eq:orth}
\end{equation}
where $\widetilde{Q}^\ell$ has orthonormal columns.

\vspace{-10pt}

\paragraph{Projection surgery on the next operator.}
Let $W^{\ell+1}\in\mathbb{R}^{m_\ell\times d_\ell}$ be the matrix form of the weight tensor of $T_{\ell+1}$ in the edit space of stage $\ell$.
We remove sensitivity to the subspace spanned by the forget directions via
\begin{equation}
W^{\ell+1'}
\;=\;
W^{\ell+1}
\Big(I-\alpha_\ell\,\widetilde{Q}^{\ell}{\left(\widetilde{Q}^{\ell}\right)}^{\top}\Big).
\label{eq:proj_update_multi}
\end{equation}
This projection-style update is aligned with prior subspace-editing approaches that suppress targeted directions while preserving the remaining representational space~\cite{cai2024edit,uppaal2025modeleditingrobustdenoised,biswas2025cure}.
We keep the bias $b^{\ell+1}$ unchanged.
For operators whose weights are stored as tensors, we apply Eq.~\eqref{eq:proj_update_multi} to the flattened matrix $W^{\ell+1}$ and reshape back.

\vspace{-10pt}

\paragraph{Layer coefficient from probe separability and depth.}
The magnitude is controlled by $\alpha_\ell\in[0,1]$, shared across all forget classes at stage $\ell$.
We set $\alpha_\ell$ using a scalar probe that measures how well stage $\ell$ separates the union of forget classes from retain.
Let $h^\ell(x)$ be a fixed-dimensional pooled summary of the stage-$\ell$ activation (e.g., global average pooling when the activation is a tensor).
Following the standard use of linear probes to assess the separability of intermediate representations~\cite{alain2016understanding}, we train a logistic classifier with labels $y=1$ for $x\in\mathcal{D}_{\mathcal{F}}$ and $y=0$ for $x\in\mathcal{D}_r$:
\begin{equation}
p_\ell(y{=}1\mid x)=\sigma\big(w_\ell^\top h^\ell(x)+b_\ell\big),
\end{equation}
and denote its held-out accuracy by $a_\ell\in[0,1]$.

We convert $a_\ell$ to a probe-based strength with chance baseline $0.5$:
\begin{equation}
\alpha_\ell^{\mathrm{probe}}
=
\min\Big\{1,\max\{0,2a_\ell-1\}\Big\}.
\label{eq:alpha_probe}
\end{equation}
To enforce small edits in early stages and larger edits in deeper stages, we apply a deterministic depth ramp:
\begin{equation}
\alpha_\ell^{\mathrm{depth}}=\frac{\ell}{L},
\label{eq:alpha_depth}
\end{equation}
and set
\begin{equation}
\alpha_\ell=\alpha_\ell^{\mathrm{probe}}\cdot \alpha_\ell^{\mathrm{depth}}.
\label{eq:alpha_final}
\end{equation}
Eq.~\eqref{eq:alpha_probe}--\eqref{eq:alpha_final} introduce no tunable hyperparameters. More analysis of $\alpha$ in~\ref{sec:supp_alpha}.

\vspace{-10pt}

\paragraph{One-shot editing over all stages.}
We compute $\{\boldsymbol{\mu}_c^\ell\}$, $\{\boldsymbol{\mu}_f^\ell\}_{f\in\mathcal{F}}$, $\{\widetilde{Q}^\ell\}$, and $\{\alpha_\ell\}$ once from the pretrained model and apply Eq.~\eqref{eq:proj_update_multi} sequentially for $\ell=L,\dots,1$ (deep to shallow), producing $\theta'$.
The overall procedure is therefore post-hoc, gradient-free, and one-shot, matching the efficiency goal of retrain-free class-unlearning methods while operating through structured weight editing rather than iterative optimization~\cite{kodge2024deep,meng2022locating,meng2022mass}.

%% file: sec/4_experimants.tex
\section{Experimental Setup}
\label{sec:exp}

\paragraph{Datasets and forgetting protocol.}
We evaluate class unlearning on four image-classification benchmarks with increasing visual complexity and label cardinality: MNIST~\cite{lecun2002gradient}, CIFAR-10, CIFAR-100~\cite{krizhevsky2009learning}, and Tiny ImageNet~\cite{wu2017tiny}. For each run, we specify a forget-class set $\mathcal{F}$ and treat all remaining classes as retain classes, following the standard class-unlearning setting studied in prior work~\cite{kodge2024deep}. We consider both \textbf{single-class} and \textbf{multi-class} forgetting. For MNIST and CIFAR-10, single-class forgetting consists of five runs, each forgetting one class, while multi-class forgetting consists of one run with two randomly selected forget classes. For CIFAR-100 and Tiny ImageNet, single-class forgetting again consists of five runs, each forgetting one class, while multi-class forgetting consists of one run in which $5\%$ of classes are randomly selected as forget classes. For each dataset-architecture pair, the same forget-set configuration is used across all compared methods. Results for single-class forgetting are averaged over the five runs unless otherwise noted.

\vspace{-10pt}

\paragraph{Models and stage definition.}
We evaluate three architectures: 
% CNN-5, 
a custom 5-layer convolutional network~(CNN-5)~\ref{supp:CNN}, ResNet-18~\cite{he2016deep}, and ViT~\cite{dosovitskiy2020image}. Each model is first trained on the full training set to obtain a pretrained baseline prior to unlearning. To apply \methodname{} consistently across architectures, we expose each network through a unified $L=5$ stage abstraction. For CNN-5, the five convolutional blocks define the stages. For ResNet-18, the stages correspond to the stem followed by the four residual groups. For ViT, the transformer blocks are partitioned into five ordered groups. At each stage, \methodname{} edits the next learnable operator, as defined in Sec.~\ref{sec:method}. 
% Detailed \methodname{} Pseudocode in~\ref{alg:damp}.
Detailed \methodname{} pseudocode is provided in Alg.~\ref{alg:damp}.

\begin{table*}[t]
\centering
\caption{Machine unlearning results for \textbf{single-class} and \textbf{multi-class} forgetting scenarios. We report retain accuracy ($\mathcal{R}{\text{acc}}$) and forget accuracy ($\mathcal{F}{\text{acc}}$), where higher $\mathcal{R}{\text{acc}}$ and lower $\mathcal{F}{\text{acc}}$ are better. Our approach mostly matches prior methods, while achieving selectivity closer to that of the retrained network, showing that \methodname{} can erase forget-class representations from hidden layers while remaining competitive with other methods in terms of both forget-class and retain-class accuracy.}
\vspace{-5pt}
\label{tab:combined}
\setlength{\tabcolsep}{4pt}
\renewcommand{\arraystretch}{1.2}
\resizebox{0.95\textwidth}{!}{
\small
\begin{tabular}{ll cccccccc cccccccc}
\toprule
& & \multicolumn{8}{c}{\textbf{Single-Class}} & \multicolumn{8}{c}{\textbf{Multi-Class}} \\
\cmidrule(lr){3-10} \cmidrule(lr){11-18}
& & \multicolumn{2}{c}{\textbf{MNIST}} & \multicolumn{2}{c}{\textbf{CIFAR-10}} & \multicolumn{2}{c}{\textbf{CIFAR-100}} & \multicolumn{2}{c}{\textbf{Tiny-ImageNet}}
  & \multicolumn{2}{c}{\textbf{MNIST}} & \multicolumn{2}{c}{\textbf{CIFAR-10}} & \multicolumn{2}{c}{\textbf{CIFAR-100}} & \multicolumn{2}{c}{\textbf{Tiny-ImageNet}} \\
\cmidrule(lr){3-4}   \cmidrule(lr){5-6}   \cmidrule(lr){7-8}   \cmidrule(lr){9-10}
\cmidrule(lr){11-12} \cmidrule(lr){13-14} \cmidrule(lr){15-16} \cmidrule(lr){17-18}
\textbf{Arch.} & \textbf{Method}
  & $\mathcal{R}_{\text{acc}}$ & $\mathcal{F}_{\text{acc}}$
  & $\mathcal{R}_{\text{acc}}$ & $\mathcal{F}_{\text{acc}}$
  & $\mathcal{R}_{\text{acc}}$ & $\mathcal{F}_{\text{acc}}$
  & $\mathcal{R}_{\text{acc}}$ & $\mathcal{F}_{\text{acc}}$
  & $\mathcal{R}_{\text{acc}}$ & $\mathcal{F}_{\text{acc}}$
  & $\mathcal{R}_{\text{acc}}$ & $\mathcal{F}_{\text{acc}}$
  & $\mathcal{R}_{\text{acc}}$ & $\mathcal{F}_{\text{acc}}$
  & $\mathcal{R}_{\text{acc}}$ & $\mathcal{F}_{\text{acc}}$ \\
\midrule
\rowcolor{rowgray}
\multirow{10}{*}{\rotatebox{90}{\textbf{CNN-5}}}
    \cellcolor{white} & Baseline        & 99.55 & 99.78 & 89.60 & 82.50 & 67.53 & 44.00 & 54.92 & 26.00
                   & 99.56 & 99.59 & 89.66 & 87.10 & 67.57 & 58.33 & 54.78 & 54.67 \\
\rowcolor{rowgray}
    \cellcolor{white} & Retrained         & 99.51 &  0.00 & 90.70 &  0.00 & 67.86 &  0.00 & 54.85 &  0.00
                   & 99.65 &  0.00 & 92.94 &  0.00 & 68.13 &  0.00 & 54.69 &  0.00 \\
  & GAU~\cite{mavrothalassitis2025ascent}
                   & 83.28 &  0.00 & 81.58 &  0.00 & 59.17 &  0.00 & 49.68 &  0.00
                   & 67.19 &  0.00 & 76.70 &  0.07 &  7.38 &  0.00 & 19.83 &  0.00 \\
  & KDU~\cite{bonato2024retain}
                   & 99.40 & 87.67 & 89.34 & 35.50 & 65.29 & 15.00 & 49.98 &  8.00
                   & 99.59 & 82.71 & 91.76 & 48.67 & 65.13 & 29.00 & 50.57 & 14.00 \\
  & DD-FT~\cite{chourasia2023forget}
                   & 99.60 &  0.00 & 89.43 &  0.00 & 64.65 &  0.00 & 50.78 &  0.00
                   & 99.52 &  0.00 & 92.54 &  0.00 & 64.97 &  0.00 & 50.69 &  0.00 \\
  & LM~\cite{baumhauer2022machine}
                   & 99.63 &  0.00 & 90.90 &  0.00 & 67.71 &  0.00 & 54.97 &  0.00
                   & 99.69 &  0.00 & 93.04 &  0.00 & 67.96 &  0.00 & 54.92 &  0.00 \\
  & RandRelabel~\cite{li2023random}
                   & 99.49 &  0.00 & 90.39 &  0.00 & 67.08 &  0.00 & 53.74 &  0.00
                   & 99.70 &  0.00 & 92.43 &  0.00 & 66.85 &  0.33 & 53.19 &  0.00 \\
  & SSD~\cite{foster2024fast}
                   & 99.55 & 0.00 & 94.09 &  0.00
                   & 66.18 &  0.00 & 54.80 &  0.00
                   & 99.56 &  0.00 & 89.66 &  0.00 & 44.71 &  0.00 & 54.94 &  0.00 \\
  & SalUn~\cite{fan2023salun}
                   & 99.60 & 0.00 & 90.63 &  0.00
                   & 67.80 &  0.00 & 54.20 &  0.00
                   & 99.80 &  0.00 & 92.53 &  0.00 & 67.45 &  1.00 & 54.39 &  0.00 \\
  & \textbf{DAMP (Ours)}
                   & 99.47 &  0.00 & 90.33 &  0.00 & 67.35 &  0.00 & 50.38 &  0.00
                   & 99.37 &  0.00 & 90.34 &  0.00 & 67.42 &  0.00 & 34.68 &  0.00 \\
\midrule
\rowcolor{rowgray}
\multirow{10}{*}{\rotatebox{90}{\textbf{ResNet-18}}}
    \cellcolor{white} & Baseline        & 99.42 & 99.33 & 84.80 & 75.60 & 59.65 & 38.00 & 53.44 & 42.00
                   & 99.46 & 99.28 & 84.61 & 82.17 & 59.64 & 52.67 & 53.34 & 56.00 \\
\rowcolor{rowgray}
    \cellcolor{white} & Retrained         & 99.42 &  0.00 & 85.96 &  0.00 & 59.05 &  0.00 & 53.12 &  0.00
                   & 99.72 &  0.00 & 87.96 &  0.00 & 59.55 &  0.00 & 53.15 &  0.00 \\
  & GAU~\cite{mavrothalassitis2025ascent}
                   & 28.23 &  0.00 & 59.08 &  0.00 &  3.70 &  0.00 &  7.15 &  0.00
                   & 40.83 &  0.00 & 49.31 &  0.00 &  1.19 &  0.00 &  0.51 &  0.00 \\
  & KDU~\cite{bonato2024retain}
                   & 99.09 & 86.77 & 83.79 & 36.40 & 55.08 & 11.00 & 47.37 & 16.00
                   & 99.17 & 92.44 & 87.10 & 62.73 & 53.24 & 25.00 & 49.21 & 28.00 \\
  & DD-FT~\cite{chourasia2023forget}
                   & 99.30 &  0.00 & 85.44 &  0.00 & 55.78 &  0.00 & 48.99 &  0.00
                   & 99.59 &  0.00 & 88.21 &  0.00 & 56.85 &  0.00 & 48.91 &  0.00 \\
  & LM~\cite{baumhauer2022machine}
                   & 99.47 &  0.00 & 86.82 &  0.00 & 59.76 &  0.00 & 53.49 &  0.00
                   & 99.58 &  0.00 & 89.36 &  0.00 & 60.01 &  0.00 & 53.45 &  0.00 \\
  & RandRelabel~\cite{li2023random}
                   & 99.54 &  0.00 & 86.32 &  0.00 & 58.48 &  0.00 & 50.67 &  0.00
                   & 99.61 &  0.00 & 88.51 &  0.00 & 58.46 &  0.00 & 51.29 &  0.00 \\
  & SSD~\cite{foster2024fast}
                   & 99.07 & 0.00 & 65.11 &  0.00
                   & 31.90 &  0.00 & 25.51 &  0.00
                   & 99.43 &  0.00 & 66.31 &  0.00 & 30.51 &  0.00 & 25.55 &  0.00 \\
  & SalUn~\cite{fan2023salun}
                   & 99.22 & 0.00 & 63.96 &  0.00
                   & 30.12 &  0.00 & 26.15 &  0.00
                   & 99.23 &  0.00 & 69.07 &  0.00 & 30.63 &  0.00 & 25.82 &  0.00 \\
  & \textbf{DAMP (Ours)}
                   & 99.70 &  0.00 & 86.20 &  0.00 & 59.34 &  0.00 & 52.80 &  0.00
                   & 99.63 &  0.00 & 83.92 &  0.00 & 60.84 &  0.00 & 52.12 &  0.00 \\
\midrule
\rowcolor{rowgray}
\multirow{10}{*}{\rotatebox{90}{\textbf{ViT}}}
    \cellcolor{white} & Baseline        & 99.02 & 98.88 & 73.89 & 76.50 & 48.11 & 27.00 & 37.60 & 20.00
                   & 98.89 & 99.31 & 74.44 & 73.47 & 48.03 & 43.67 & 37.48 & 39.33 \\
\rowcolor{rowgray}
    \cellcolor{white} & Retrained         & 98.95 &  0.00 & 76.88 &  0.00 & 50.18 &  0.00 & 37.57 &  0.00
                   & 98.77 &  0.00 & 79.10 &  0.00 & 46.80 &  0.00 & 37.07 &  0.00 \\
  & GAU~\cite{mavrothalassitis2025ascent}
                   & 99.33 &  0.00 & 80.29 &  0.00 & 48.97 &  0.00 & 35.13 &  0.00
                   & 99.11 &  0.00 & 83.46 &  0.00 & 47.46 &  0.00 & 34.57 &  0.00 \\
  & KDU~\cite{bonato2024retain}
                   & 99.10 & 23.88 & 74.32 & 56.30 & 48.94 &  0.00 & 37.90 &  0.00
                   & 99.13 & 22.19 & 76.16 & 55.67 & 49.60 &  2.00 & 38.71 &  6.00 \\
  & DD-FT~\cite{chourasia2023forget}
                   & 99.20 &  0.00 & 79.44 &  0.00 & 49.88 &  0.00 & 36.63 &  0.00
                   & 99.44 &  0.00 & 83.29 &  0.00 & 50.28 &  0.00 & 36.86 &  0.00 \\
  & LM~\cite{baumhauer2022machine}
                   & 99.10 &  0.00 & 77.81 &  0.00 & 48.16 &  0.00 & 37.66 &  0.00
                   & 99.14 &  0.00 & 82.49 &  0.00 & 48.25 &  0.00 & 37.63 &  0.00 \\
  & RandRelabel~\cite{li2023random}
                   & 99.53 &  0.00 & 82.61 &  0.00 & 52.11 &  0.00 & 39.09 &  0.00
                   & 99.54 &  0.00 & 85.61 &  0.00 & 51.87 &  0.00 & 38.98 &  0.00 \\
  & SSD~\cite{foster2024fast}
                   & 99.07 & 0.00 & 75.39 &  0.00
                   & 45.91 &  0.00 & 37.71 &  0.00
                   & 98.89 &  0.00 & 76.39 &  0.00 & 48.04 &  0.00 & 37.67 &  0.00 \\
  & SalUn~\cite{fan2023salun}
                   & 99.57 & 0.00 & 82.48 &  0.00
                   & 51.95 &  0.00 & 39.67 &  0.00
                   & 99.32 &  0.00 & 85.21 &  0.00 & 51.48 &  0.67 & 39.26 &  0.00 \\
  & \textbf{DAMP (Ours)}
                   & 99.18 &  0.00 & 76.47 &  0.00 & 48.02 &  0.00 & 37.51 &  0.00
                   & 99.17 &  0.00 & 80.16 &  0.00 & 47.96 &  0.00 & 37.40 &  0.00 \\
\bottomrule
\end{tabular}
}
\vspace{-15pt}
\end{table*}

\vspace{-10pt}

\paragraph{Compared methods.}
We compare against the following baselines: (i) Baseline, the original model trained on the full dataset before unlearning; (ii) Retrain, retraining from scratch on retain-only data, used as the gold-standard \textit{as-if-never-seen} reference; (iii) GAU~\cite{mavrothalassitis2025ascent}, gradient-ascent unlearning, which increases forget loss while encouraging retention of non-forget performance; (iv) KDU~\cite{bonato2024retain}, knowledge-distillation unlearning, which distills retain behavior from the original model while pushing forget outputs toward a uniform distribution; (v) DD-FT~\cite{chourasia2023forget}, classifier reinitialization followed by fine-tuning on retain-only data; (vi) LM~\cite{baumhauer2022machine}, inference-time logit masking of the forget classes; (vii) RandRelabel~\cite{li2023random}, relabeling forget examples with randomly sampled retain-class labels followed by fine-tuning; (viii) SSD~\cite{foster2024fast}, selective synaptic dampening, which suppresses weights identified as important to the forget data without full retraining; and (ix) SalUn~\cite{fan2023salun}, saliency unlearning, which uses gradient-based weight saliency to guide targeted forgetting while preserving retain-task behavior. These baselines span most of the main fine-tuning, gradient, distillation, masking, relabeling, saliency, and representation-unlearning families commonly used in class-unlearning evaluations~\cite{kodge2024deep,cadet2025deep}.

\vspace{-10pt}

\paragraph{Training and evaluation.}
All methods are evaluated from the same pretrained baseline for each dataset--architecture pair. Retrain is trained on retain-only data, and all post-hoc unlearning methods use the same retain/forget partition and matched training budget for each run. We report retain accuracy on the retain-class test split and forget accuracy on the forget-class test split~\cite{kodge2024deep,cadet2025deep}. Because output-level forgetting can reflect either representational removal or classifier-head suppression, we additionally evaluate selectivity, layer-wise representational similarity, and final-layer bias shifts (Secs.~\ref{sec:supp_selectivity}--\ref{sec:supp_bias}). Full training hyperparameters are provided in~\ref{supp:Imp}.

%% file: sec/5_result.tex
\vspace{-10pt}
\section{Results}
\label{sec:res}

We evaluate \methodname{} on standard class-unlearning benchmarks using retain accuracy and forget accuracy summarized in Table~\ref{tab:combined} and Figs.~\ref{fig:spider} and ~\ref{fig:rdm_intro}. Across single-class and multi-class settings, \methodname{} consistently achieves a strong retain-forget tradeoff and closely matches retraining. This trend holds across MNIST, CIFAR-10, CIFAR-100, and Tiny-ImageNet, and across CNN-5, ResNet-18, and ViT architectures. In contrast, KDU often leaves substantial residual forget-class accuracy, while GAU frequently reduces retain performance more aggressively. DD-FT, although it achieves performance close to the retrained network, has a higher computational cost than most of the other methods. LM performs well, but it only affects the logits of the forgotten class set, meaning it has no effect on the representations. RandRelabel also performs close to the retrained network, but it alters the geometry of the problem by redirecting the forgotten class to a random class. SSD consistently suppresses forget-class accuracy to zero, but this strong forgetting is frequently accompanied by significant degradation in retain accuracy, particularly on CIFAR-100 and Tiny-ImageNet. SalUn likewise attains near-zero forget accuracy and performs reasonably well in some CNN-5 and ViT cases, yet its retain performance is less stable across architectures and datasets, especially for ResNet-18.

We also evaluated the compared methods under adversarial perturbation attacks, as provided in Sec.~\ref{sec:supp_adv}., we  show that methods with similar output-level forgetting can differ substantially in retained-class robustness and in residual vulnerability on forgotten classes. In particular, \methodname{} maintains strong adversarial retain performance while keeping adversarial forget accuracy at zero.
% Tables~\ref{tab:single_class_grouped} and~\ref{tab:multi_class_grouped} and Figs.~\ref{fig:rdm_intro},~\ref{fig:spider} summarize these results.

% This behavior is also reflected in the t-SNE~\cite{maaten2008visualizing} visualizations in Fig.~\ref{fig:tsne}. Relative to retraining, effective unlearning should reduce the separability of the forget class without disrupting the layout of retained classes. \methodname{} produces feature layouts that are visually closer to retraining, while several baselines either preserve tighter forget-class clustering or distort retained-class structure more strongly. Although t-SNE is only qualitative, it supports the quantitative evidence from selectivity and RDM analysis.

\subsection{Detailed Analyses}

% We analyze whether apparent forgetting is driven by representational removal or by suppression at the classifier head. Fig.~\ref{fig:bias_related} shows that several baselines shift the final-layer bias of the forget class strongly negative, which can reduce forget accuracy without removing deeper forget-class evidence. The supplementary bias analysis further confirms that output suppression alone can produce the appearance of forgetting. In contrast, \methodname{} relies less on this mechanism, consistent with its stronger selectivity and representation-level behavior.

Fig.~\ref{fig:alpha} studies the effect of the depth-aware scaling factor $\alpha$, which controls the strength of the projection applied at each layer. The figure shows that this scaling is not a cosmetic design choice: weaker projection leaves substantial residual forget performance, whereas stronger projection increases damage to retained classes. The dynamic setting yields a better operating point between these two failure modes, achieving zero forget accuracy without the corresponding loss in retain accuracy seen at more aggressive settings. This result supports the main design of \methodname{} by showing that layer-dependent projection strength is necessary to balance forgetting against preservation, rather than treating all layers identically.

Fig.~\ref{fig:biassweep} isolates a different failure mode. By explicitly sweeping the final-layer bias of the forget class,  forget accuracy can be reduced substantially through output-level suppression alone. However, this reduction does not imply that the model has removed forget-class information from its internal representations; it only makes the class less likely to be predicted at the classifier head. As discussed in~\cite{hinton2015distilling}, when the network sees one class less often than the others, it starts to reduce the bias for that class without changing the feature extraction weights. 

% Moreover, as shown in~\ref{fig:tsne}, the retrained network does not learn the high-level features of the forget class. The problem with other methods is that they not only reduce the bias but also retain information about the forget class.

Table~\ref{tab:continuous_unlearning_main} evaluates continual unlearning, where classes are forgotten sequentially over multiple rounds rather than in a single step. This setting is more demanding because errors can accumulate: a method may leave residual information from earlier forget classes, or progressively damage retained performance after repeated updates. The table shows that \methodname{} remains close to retraining across rounds, maintaining high retain accuracy while keeping both newly forget accuracy (NF) and all-forget accuracy (AF) near zero.

Finally, Fig.~\ref{fig:seg_qual} extends the evaluation beyond classification and tests whether the same unlearning behavior appears in dense prediction. The segmentation examples compare the baseline model, retraining, \methodname{}, and competing baselines on representative images. The key question is not only whether the forgotten category is suppressed, but whether this can be done without degrading surrounding retained regions. In these examples, \methodname{} produces outputs that are visually correct: forgotten regions are removed more cleanly, while the spatial structure and semantic coherence of retained classes are better preserved.

% For completeness, the full DAMP procedure is provided in Algorithms~\ref{alg:damp} in the supplementary material. These algorithms specify the computation of class-wise edit-space prototypes, the estimation of layer-wise coefficients $\alpha_\ell$, and the final projection step used to remove forget-specific directions from the model weights.

% Beyond achieving competitive retain and forget accuracy, \methodname{} improves selective forgetting, reduces residual forget-class structure in deep layers, and avoids relying primarily on classifier-head suppression. Together, these results support representation-aware unlearning as a stronger criterion than output-level forgetting alone.

%% file: sec/6_limit.tex
\vspace{-7pt}
\section{Limitations}
\label{sec:limit}

The current formulation of our method has several limitations, which may need to be addressed for other application settings. In particular, under our setting, the representation of a class can be approximated largely by its mean feature vector, and class representations are assumed to be, mostly, linearly separable. These assumptions may not hold for highly multimodal, anisotropic distributions. Our framework could be extended to address this by expanding the computations to estimate class-specific subspaces that account for covariance-sensitive directions. Importantly, the present formulation remains a practical and effective starting point because, in deeper layers, class information often concentrates in a relatively low-dimensional set of directions. Hence, our formulation is an efficient and stable approach for many class-erasure scenarios, preserving shared features among retained classes, while also being extensible to cases where the mean feature vector alone is insufficient.

%% file: sec/7_conclusion.tex
\section{Conclusion}
\label{sec:con}

This work argues that class unlearning should be evaluated beyond output accuracy, since apparent forgetting can arise from classifier-head suppression while forget-class evidence remains encoded in deep representations. We introduced \methodname{}, a gradient-free, one-shot, depth-aware projection method that removes forget-specific directions from pretrained networks. Across multiple benchmarks and architectures, \methodname{} more closely approximates retraining than several prior baselines, improving selective forgetting, preserving retain-class utility, and reducing residual forget-class structure in deep layers. These findings suggest that structured representation editing is a promising direction for scalable and more faithful machine unlearning, while future work should extend the method to settings with more complex, multimodal class geometry.

\section*{Acknowledgments}

The authors thank their affiliated institutions for support. Arman Hatami (AH) and Ilya E.\ Monosov (IEM) were supported by the National Institute of Mental Health (NIMH) under grant R01 MH128344. Romina Aalishah (RA) was supported by the Johns Hopkins Department of Electrical and Computer Engineering. 

AH led the project and developed DAMP. RA contributed to the development and testing of DAMP and participated in writing the manuscript. IEM advised the development of DAMP, assisted with manuscript preparation, and secured funding for the project.

We are grateful to all members of the Laboratory of Adaptive and Maladaptive Intelligence (LAMI) for the helpful
discussions that improved this manuscript.

%% file: sec/X_suppl.tex
\appendix
\renewcommand{\thefigure}{S\arabic{figure}}
\renewcommand{\thetable}{S\arabic{table}}
\renewcommand{\thealgorithm}{S\arabic{algorithm}}
\setcounter{table}{0}
\setcounter{page}{1}
\maketitlesupplementary

\section{Additional Analyses and Supplementary Results}
\label{sec:supplementary}

\subsection{Implementation detail}
\label{supp:Imp}
All experiments were run on a single NVIDIA A100 GPU with 80\,GB memory. We used PyTorch with a fixed random seed of 42 for Python, NumPy, and CUDA; CuDNN was set to deterministic mode and benchmarking was disabled. Data loading used 4 workers. For baseline training and retraining, CNN-5, ResNet-18 were optimized with SGD with momentum 0.9. On MNIST, these models were trained for 30 epochs with learning rate 0.01 and weight decay $10^{-4}$, without cosine scheduling. On CIFAR-10, CIFAR-100, and Tiny ImageNet, they were trained for 50 epochs with learning rate 0.1 and weight decay $5\times10^{-4}$, using cosine annealing. ViT models were optimized with AdamW. On MNIST, ViT used learning rate $10^{-3}$, weight decay 0.05, and 30 epochs. On CIFAR-10 and CIFAR-500, ViT used learning rate $5\times10^{-4}$, weight decay 0.05, and 100 epochs. On Tiny ImageNet, ViT used learning rate $3\times10^{-4}$, weight decay 0.05, and 200 epochs. The training batch size was 128 and the evaluation batch size was 256 for all datasets.

For unlearning baselines, GAU was run for 10 epochs using Adam with learning rate $10^{-4}$ and loss $\mathcal{L}_{\mathrm{GAU}}=\mathrm{CE}(\mathrm{retain})-\lambda\,\mathrm{CE}(\mathrm{forget})$, with $\lambda=0.1$. KDU was run for 10 epochs using Adam with learning rate $10^{-4}$, temperature $T=4.0$, and forget-loss weight $\lambda=0.5$; it matched the teacher on retain samples via KL divergence and pushed forget samples toward the uniform distribution. DD-FT reinitialized the final classifier layer and fine-tuned the full network on retain data for 10 epochs using Adam with learning rate $5\times10^{-4}$. RandRelabel was run for 10 epochs using Adam with learning rate $10^{-4}$, where forget-class labels were randomly reassigned to retain classes during training. LM was implemented as inference-time logit masking by setting the logits of forget classes to $-\infty$, without any weight updates. SSD was implemented as Selective Synaptic Dampening by estimating the diagonal Fisher information on the full training set and on the forget set. Parameters satisfying $F_{\mathrm{forget}}/F_{\mathrm{train}} > \alpha$ were selected with $\alpha = 25.0$, and selected parameters were dampened as
$\theta \leftarrow \theta / \bigl(1 + \lambda\, F_{\mathrm{forget}}/F_{\mathrm{train}}\bigr)$
with $\lambda = 1.0$. SSD used no optimizer-based fine-tuning or weight-update epochs.

SalUn was run for 10 epochs using Adam with learning rate $10^{-4}$. A saliency mask was first computed from the absolute gradients of the forget-set loss, and the top $50\%$ most salient weights were retained (saliency threshold $\tau = 0.5$). During fine-tuning, forget-class labels were randomly reassigned to retain classes, and gradients on non-salient weights were zeroed so that only salient weights were updated. For all methods, hyperparameters were adjusted for certain datasets and architectures to achieve the best balance between retention and forgetting accuracy.

\paragraph{\methodname{} details.}
Class-wise edit means were computed from clean training data. For each layer, forget-class directions were obtained as residuals after projection onto the span of retain-class means, with residual threshold $\epsilon=10^{-8}$. Multiple forget directions were orthonormalized with QR decomposition. The layer-wise edit strength was set as $\alpha_\ell=\alpha_{\mathrm{probe},\ell}\cdot \alpha_{\mathrm{depth},\ell}$, where $\alpha_{\mathrm{probe},\ell}=\mathrm{clip}(2a_\ell-1,0,1)$ was derived from the accuracy $a_\ell$ of a logistic-regression probe trained on GAP features, and $\alpha_{\mathrm{depth},\ell}=(\ell+1)/5$; since $L = 5$. Edits were applied in reverse layer order using the closed-form update $W' = W(I-\alpha_\ell QQ^\top)$. Logistic-regression probes used scikit-learn with \texttt{lbfgs}, \texttt{class\_weight="balanced"}, $C=1.0$, \texttt{max\_iter}=1000, and an 80/20 train/test split after feature standardization. In some experiments, $\alpha$ was manually increased by a fixed constant to amplify the effect of weight surgery. For example, for Tiny ImageNet with ViT, all $\alpha$ values were increased by 3.0. The dynamic layer-wise scaling was still preserved, since earlier layers should be edited less than deeper layers.

\subsection{Custom CNN-5 architecture.}
\label{supp:CNN}
For all datasets, we used the same lightweight 5-stage convolutional network, denoted \textit{CNN-5}. The network supports variable input channels and numbers of output classes, and is instantiated with $1$ input channel for MNIST and $3$ input channels for CIFAR-10, CIFAR-100, and Tiny ImageNet. Each stage consists of a convolution, batch normalization, and ReLU activation, with max-pooling applied in the first three stages. The architecture is:
\[
\begin{array}{l}
\textbf{Stage 1:}\ \mathrm{Conv}(C_{\mathrm{in}},64,3\!\times\!3)\to \mathrm{BN}\to \mathrm{ReLU}\to \mathrm{Pool}\\
\textbf{Stage 2:}\ \mathrm{Conv}(64,128,3\!\times\!3)\to \mathrm{BN}\to \mathrm{ReLU}\to \mathrm{Pool}\\
\textbf{Stage 3:}\ \mathrm{Conv}(128,256,3\!\times\!3)\to \mathrm{BN}\to \mathrm{ReLU}\to \mathrm{Pool}\\
\textbf{Stage 4:}\ \mathrm{Conv}(256,256,3\!\times\!3)\to \mathrm{BN}\to \mathrm{ReLU}\\
\textbf{Stage 5:}\ \mathrm{Conv}(256,128,3\!\times\!3)\to \mathrm{BN}\to \mathrm{ReLU}
\end{array}
\]
All convolutions use padding 1, and \(\mathrm{Pool}\) denotes \(\mathrm{MaxPool}(2)\). The classifier head is
\[
\mathrm{AdaptiveAvgPool}(1)\to \mathrm{Flatten}\to \mathrm{Linear}(128,n_c),
\]
where $n_c$ is the number of classes.

\subsection{Layer-wise Selectivity Metric}
\label{sec:supp_selectivity}

To quantify the tradeoff between forget-class removal and retain-class preservation, we define a layer-wise \emph{selectivity} score. At layer $\ell$, selectivity is computed as
\begin{equation}
\begin{aligned}
\mathrm{Selectivity}_{\ell}
&=
\left(
\mathrm{AUC}^{\mathrm{baseline}}_{\mathrm{forget},\ell}
-
\mathrm{AUC}^{\mathrm{method}}_{\mathrm{forget},\ell}
\right) \\
&\quad -
\left(
\mathrm{ACC}^{\mathrm{baseline}}_{\mathrm{retain},\ell}
-
\mathrm{ACC}^{\mathrm{method}}_{\mathrm{retain},\ell}
\right).
\end{aligned}
\label{eq:selectivity}
\end{equation}
The first term measures \emph{forget removal}, namely how much the method reduces forget-vs-retain linear separability relative to the baseline model. The second term measures \emph{retain damage}, namely how much retain-only classification accuracy degrades relative to the baseline. Thus, higher selectivity indicates a more desirable operating point: the method removes more linearly accessible forget information while incurring less damage to the retained representation structure.

\subsection{Unlearning Results under FGSM and PGD Evaluation}
\label{sec:supp_adv}

In the main paper, we evaluate unlearning primarily through clean retain and forget accuracy, together with deeper representational analyses. To further assess whether the compared methods remain stable under adversarial perturbations, Table~\ref{tab:fgsm_pgd_results} reports results under FGSM and PGD evaluation. Specifically, we report retain accuracy (Retain, $\uparrow$) and forget accuracy (Forget, $\downarrow$) under both attacks for each method.

These results show that \methodname{} achieves retain accuracy under both attack types that is closer to the retrained network, while keeping forget accuracy at zero, whereas other methods show vulnerability, especially in retain performance under attack.

\begin{table}[t]
\centering
\small
\setlength{\tabcolsep}{6pt}
\renewcommand{\arraystretch}{1.08}
\begin{threeparttable}
\caption{Unlearning results under FGSM and PGD evaluation. We report retain accuracy (Retain, $\uparrow$) and forget accuracy (Forget, $\downarrow$) for each method. LM performance is close to \methodname{}, but it sets the logits of the forgotten classes to $-\infty$ only during inference, meaning there is no representational unlearning.}
\label{tab:fgsm_pgd_results}
\begin{tabular}{lcccc}
\toprule
\multirow{2}{*}{Method} 
& \multicolumn{2}{c}{FGSM} 
& \multicolumn{2}{c}{PGD} \\
\cmidrule(lr){2-3} \cmidrule(lr){4-5}
& Retain & Forget & Retain & Forget \\
\midrule
Baseline    & 80.9111 & 84.0 & 64.0    & 68.2 \\
Retrained     & 80.1778 & 0.0  & 66.2111 & 0.0  \\
GAU~\cite{mavrothalassitis2025ascent}         & 25.0667 & 0.0  & 20.0222 & 0.0  \\
KDU~\cite{bonato2024retain}         & 80.2889 & 32.4 & 63.8667 & 14.6 \\
DD-FT~\cite{chourasia2023forget}       & 78.2333 & 0.0  & 59.1889 & 0.0  \\
LM~\cite{baumhauer2022machine}          & 82.0222 & 0.0  & 65.2222 & 0.0  \\
RandRelabel~\cite{li2023random} & 80.0556 & 0.0  & 61.8667 & 0.0  \\
SSD~\cite{foster2024fast}         & 81.5333 & 0.1  & 64.6444 & 0.0  \\
SalUn~\cite{fan2023salun}       & 80.1889 & 0.0  & 63.4889 & 0.0  \\
\textbf{DAMP (Ours)}        & 81.9333 & 0.0  & 65.0778 & 0.0  \\
\bottomrule
\end{tabular}
\end{threeparttable}
\end{table}

\subsection{Full Representational Dissimilarity Matrices}
\label{sec:supp_rdm}

\begin{figure*}[t]
    \centering
    \includegraphics[width=0.97\textwidth]{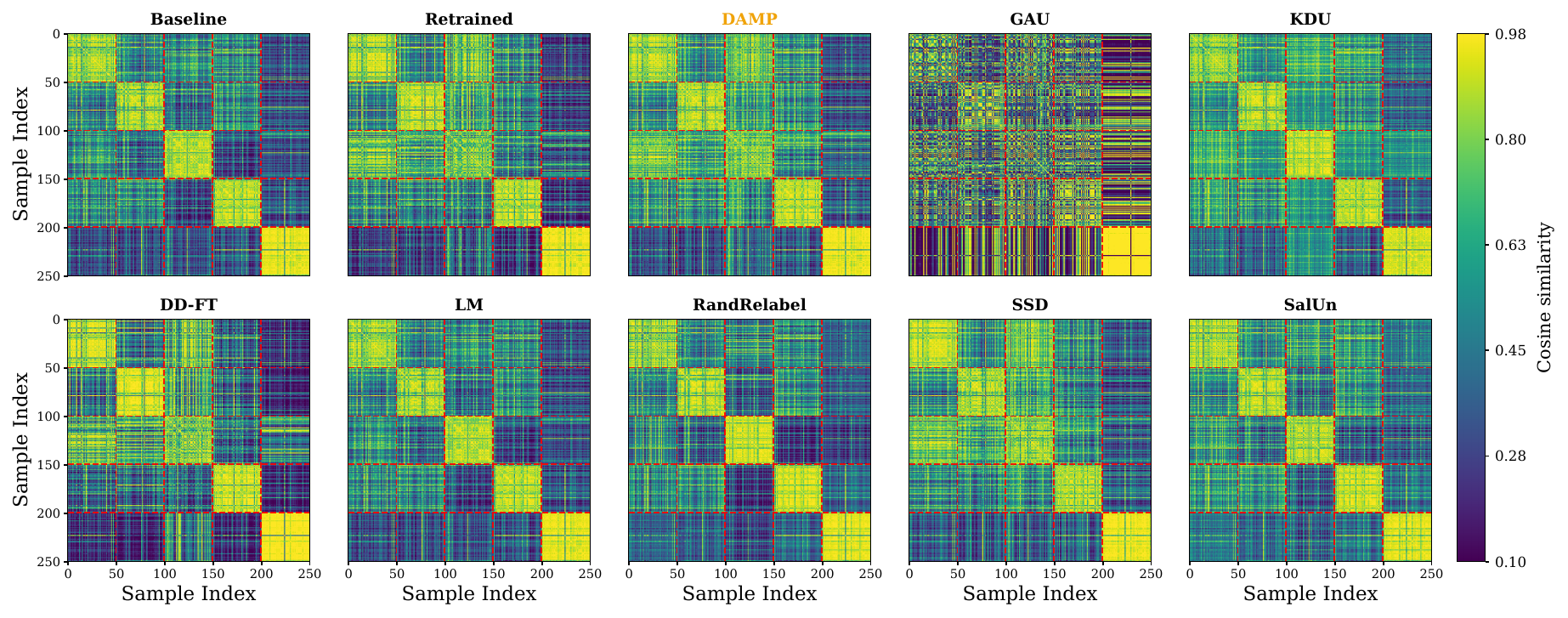}
    \caption{Full representational dissimilarity matrices (RDMs)  comparing \methodname{} with the baseline, Gradient Ascent Unlearning (GAU), knowledge-distillation unlearning (KDU), Data Deletion Fine-Tuning (DD-FT), Logit Masking (LM), Random Relabeling~(RandRelabel), Selective Synaptic Dampening~(SSD), and Saliency Unlearning~(SalUn) for a 5-layer CNN~(CNN-5) on CIFAR-10; retain classes 4 (Deer), 5 (Dog), 7 (Horse), 8 (Ship) and forget class 6 (Frog).}
    \label{fig:rdm_org}
\end{figure*}

In the main paper, Fig.~\ref{fig:rdm_intro} reports the difference between the representational dissimilarity matrix (RDM) of each method and the retrained reference model. To provide the underlying representation structure directly, Fig.~\ref{fig:rdm_org} shows the full RDMs for the compared methods at the same deep layer (layer 5). These plots offer a more complete view of how forget-class structure and retained-class geometry are organized in feature space after unlearning.

\subsection{t-SNE Visualization of Deep Representations}
\label{sec:supp_tsne}

\begin{figure*}[t]
    \centering
    \includegraphics[width=0.97\textwidth]{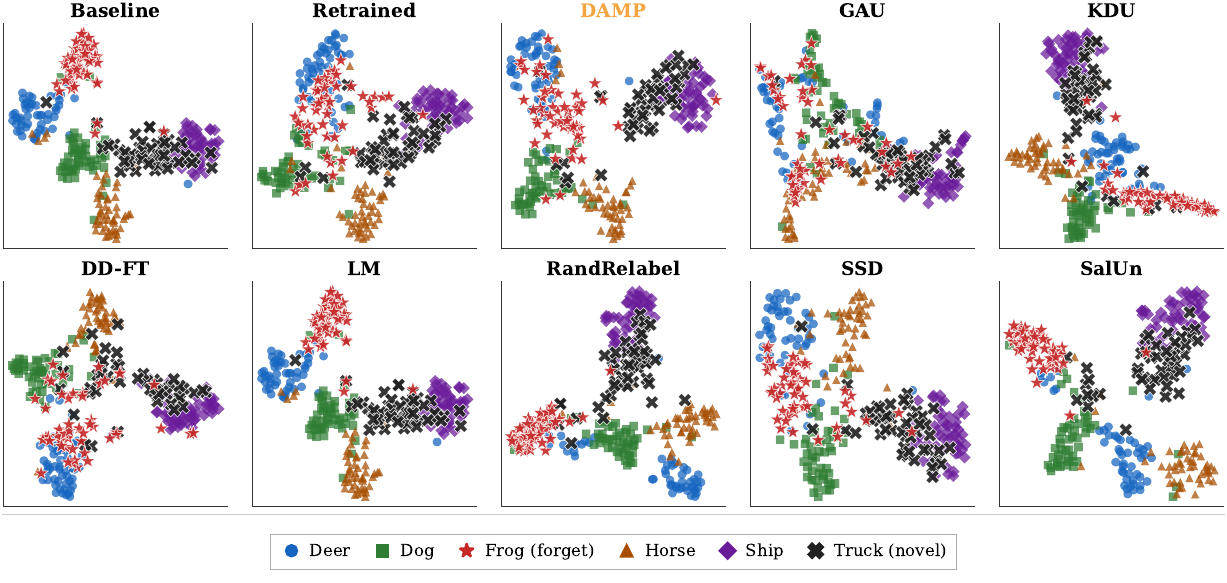}
    \caption{t-SNE visualization of deep-layer (layer 5) features after class unlearning. We project the same representations used in the RDM analysis into two dimensions for qualitative comparison across methods. The forget class is highlighted separately from retained and novel classes. \methodname{} yields a feature layout that more closely resembles retraining while reducing the visual separability of the forget class, whereas several baselines either preserve stronger forget-class clustering or introduce greater distortion among retained classes. Results are comparing \methodname{} with the baseline, Gradient Ascent Unlearning (GAU), knowledge-distillation unlearning (KDU), Data Deletion Fine-Tuning (DD-FT), Logit Masking (LM), Random Relabeling~(RandRelabel), Selective Synaptic Dampening~(SSD), and Saliency Unlearning~(SalUn) for a 5-layer CNN~(CNN-5) on CIFAR-10; retain classes 4 (Deer), 5 (Dog), 7 (Horse), 8 (Ship) and forget class 6 (Frog) and novel class 9 (Truck).}
    \label{fig:tsne}
\end{figure*}

To complement the RDM-based analysis, in Fig.~\ref{fig:tsne}, we also visualize deep-layer representations using t-SNE~\cite{maaten2008visualizing}. For each method, we extract features from the same layer used in the representational analysis and project them into two dimensions using a shared t-SNE configuration. This visualization provides an intuitive view of how forget, retain, and novel classes are arranged after unlearning.

Relative to retraining, an effective unlearning method should reduce the separability of the forget class without unnecessarily disrupting the structure of retained classes. Also, the unlearned networks should treat the novel and forget classes similarly, because the network should no longer retain high-level semantic information about the forget class. Nevertheless, the two should not be expected to behave identically: the model has already learned the low-level features of the forget class, whereas a novel class may contain low-level patterns the model has never encountered. Consequently, the network may preserve more low-level feature knowledge for the forget class than for the novel class. We emphasize that t-SNE is used only as a qualitative diagnostic, since it does not faithfully preserve global geometry. Our primary evidence remains the RDM analysis and the quantitative results in the main paper.

\subsection{Effect of Depth-Aware Scaling}
\label{sec:supp_alpha}

\begin{figure*}[t]
    \centering
    \includegraphics[width=0.97\textwidth]{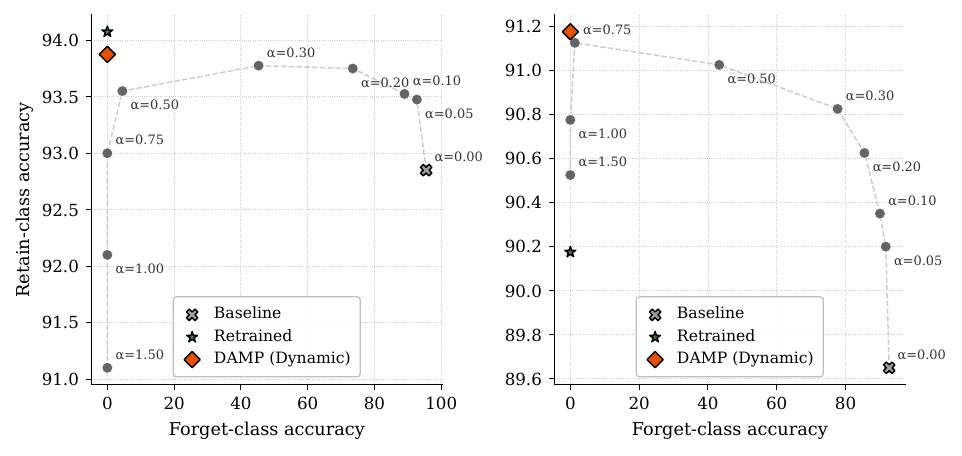}
    \caption{Left: CNN-5 Network. Right: ResNet-18 Network. Effect of the depth-aware scaling factor $\alpha$ on the retain-forget tradeoff. We vary the projection strength and report the resulting balance between retain accuracy and forget accuracy. The selected setting used in \methodname{} achieves strong forgetting while maintaining high retain performance, illustrating the benefit of depth-aware scaling. Results are shown on CIFAR-10; retain classes 4 (Deer), 5 (Dog), 7 (Horse), 8 (Ship) and forget class 6 (Frog).}
    \label{fig:alpha}
\end{figure*}

A central component of \methodname{} is the depth-aware scaling factor $\alpha$, which controls the strength of the projection across layers. Fig.~\ref{fig:alpha} presents the retain--forget tradeoff under different settings of $\alpha$. The results show that the proposed scaling yields a favorable operating point, balancing strong forgetting with preservation of retained knowledge. This analysis supports the choice used in the main experiments.

\subsection{Additional Final-Layer Bias Analysis}
\label{sec:supp_bias}

\begin{figure}[t]
    \centering
    \includegraphics[width=0.95\linewidth]{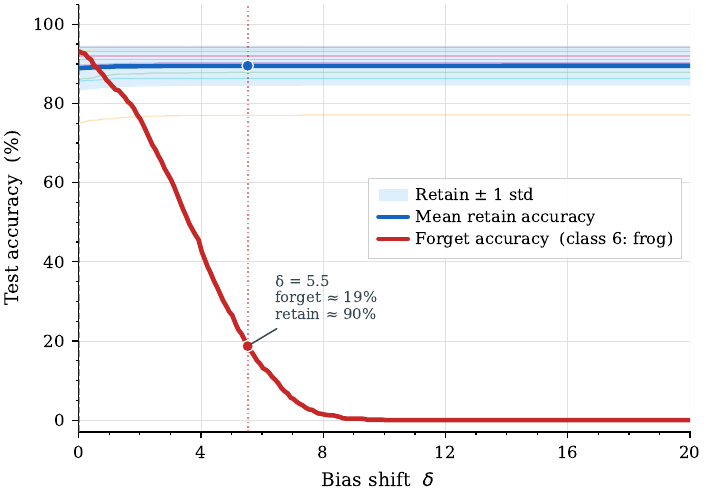}
    \caption{Additional bias-sweep analysis for the final classifier layer. We vary the bias associated with the forget class and measure the resulting change in model behavior. The results illustrate that reducing forget-class predictions can often be achieved through output-level suppression alone, reinforcing the need for representational analyses beyond classifier outputs. Results are shown for the CNN-5 architecture on CIFAR-10.}
    \label{fig:biassweep}
\end{figure}

\begin{figure}[t]
    \centering
    \includegraphics[width=0.95\linewidth]{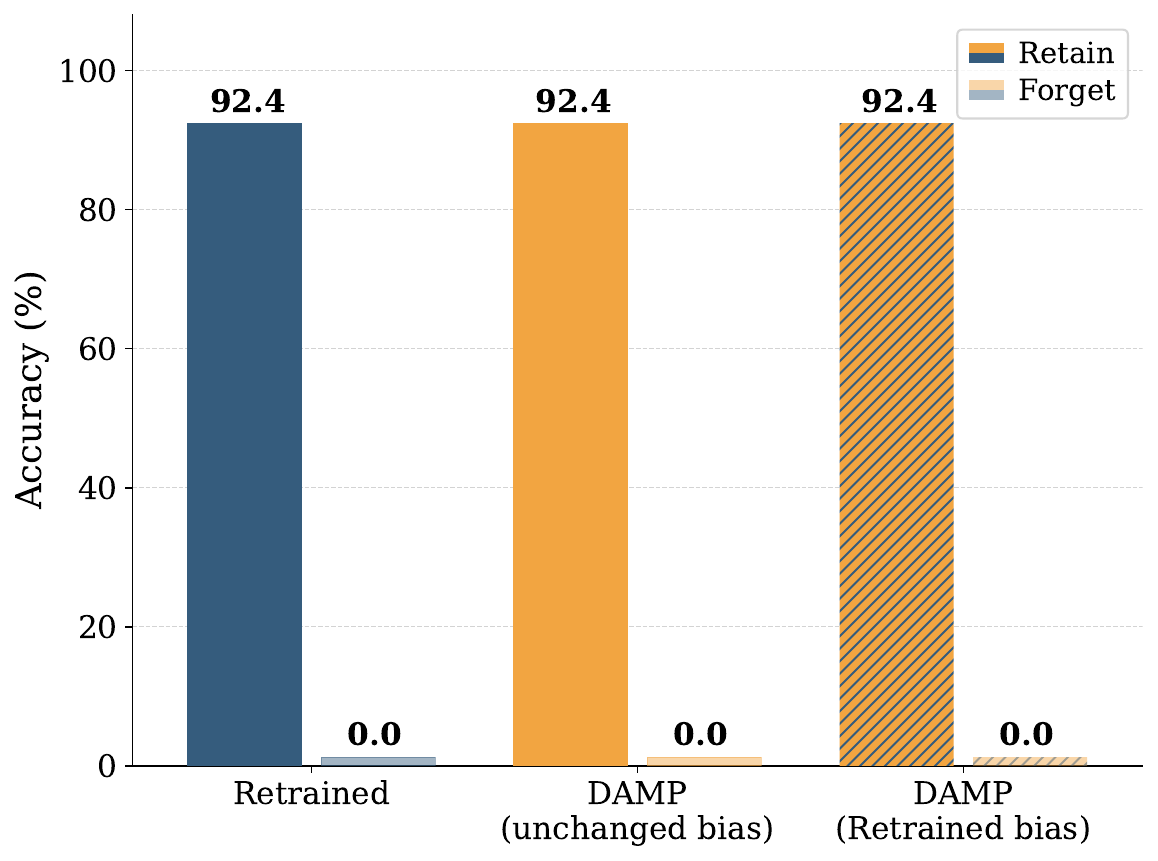}
    \caption{\methodname{} achieves forgetting through weight-space projection and representation unlearning alone. Forcing the last-layer biases to match the delta bias of the retrained network leaves retain and forget accuracy unchanged, demonstrating that bias plays no role in the forgetting mechanism for \methodname{}. Results are shown for the CNN-5 architecture on CIFAR-10.}
    \label{fig:biasdamp}
\end{figure}

The paper shows that several baselines suppress forgetting at the output level by strongly shifting the classifier bias of the forget class, rather than genuinely removing forget-class evidence from the internal representation. Fig.~\ref{fig:biassweep} extends this analysis by sweeping the final-layer bias and measuring the resulting behavior. The figure further highlights that output suppression alone can artificially reduce forget accuracy while leaving deeper representations insufficiently unlearned. However, as we show in Fig.~\ref{fig:biasdamp}, \methodname{} does not use this shortcut and achieves unlearning without relying on bias shift. Even when it has the same strong bias shift as the retrained network, the results remain unchanged.

\subsection{Continual Unlearning Results}
\label{sec:supp_continual}

Table~\ref{tab:continuous_unlearning_main} reports continual unlearning performance on CIFAR-10 across sequential class-forgetting rounds. We report retain accuracy (R), newly forget accuracy (NF), all forget accuracy (AF), and the continuous unlearning score (CUS) after each round. These results show that \methodname{} remains competitive with retraining throughout the sequential setting, while several baselines either accumulate residual forget-class information or suffer substantial degradation on retained classes.

\begin{table*}[t]
\centering
\small
\setlength{\tabcolsep}{3.2pt}
\renewcommand{\arraystretch}{1.08}
\begin{threeparttable}
\caption{Continuous unlearning results after each class-forgetting round on CIFAR-10 for the CNN-5 architecture. For each method, we report retain accuracy (R, $\uparrow$), newly forget accuracy (NF, $\downarrow$), all-forget accuracy (AF, $\downarrow$), and continuous unlearning score (CUS, $\uparrow$) after forgetting the class indicated at each round.}
\label{tab:continuous_unlearning_main}
\begin{tabular}{llccccccccc}
\toprule
\multirow{2}{*}{Method} & \multirow{2}{*}{Metric}
& \multicolumn{9}{c}{Forget class} \\
\cmidrule(lr){3-11}
&  & 1 (air.) & 2 (auto.) & 3 (bird) & 4 (cat) & 5 (deer) & 6 (dog) & 7 (frog) & 8 (horse) & 9 (ship) \\
\midrule
\multirow{4}{*}{Retrained}
& R   & 89.2 & 88.7 & 90.1 & 94.2 & 95.0 & 97.0 & 97.1 & 97.5 & 100.0 \\
& NF  & 0.0  & 0.0  & 0.0  & 0.0  & 0.0  & 0.0  & 0.0  & 0.0  & 0.0 \\
& AF  & 0.0  & 0.0  & 0.0  & 0.0  & 0.0  & 0.0  & 0.0  & 0.0  & 0.0 \\
& CUS & 89.2 & 88.7 & 90.1 & 94.2 & 95.0 & 97.0 & 97.1 & 97.5 & 100.0 \\
\midrule
\multirow{4}{*}{GAU~\cite{mavrothalassitis2025ascent}}
& R   & 43.5 & 19.2 & 30.3 & 30.8 & 37.5 & 40.9 & 33.4 & 49.2 & 1.5 \\
& NF  & 0.0  & 0.0  & 0.0  & 0.0  & 0.0  & 19.4 & 32.3 & 0.0  & 49.4 \\
& AF  & 0.0  & 0.0  & 0.0  & 0.0  & 0.0  & 3.4  & 6.6  & 13.7 & 17.6 \\
& CUS & 43.5 & 19.2 & 30.3 & 30.8 & 37.5 & 32.9 & 22.6 & 49.2 & 0.8 \\
\midrule
\multirow{4}{*}{KDU~\cite{bonato2024retain}}
& R   & 87.6 & 85.8 & 88.8 & 91.3 & 91.3 & 92.2 & 92.8 & 93.9 & 97.0 \\
& NF  & 52.3 & 60.1 & 47.0 & 38.3 & 44.1 & 24.6 & 31.0 & 44.8 & 59.8 \\
& AF  & 52.3 & 56.3 & 59.6 & 60.7 & 60.8 & 52.5 & 41.8 & 46.4 & 53.3 \\
& CUS & 41.8 & 34.2 & 47.0 & 56.3 & 51.0 & 69.5 & 64.0 & 51.8 & 39.0 \\
\midrule
\multirow{4}{*}{DD-FT~\cite{chourasia2023forget}}
& R   & 88.6 & 88.5 & 90.3 & 93.9 & 96.1 & 97.6 & 98.2 & 98.0 & 100.0 \\
& NF  & 0.0  & 0.0  & 0.0  & 0.0  & 0.0  & 0.0  & 0.0  & 0.0  & 0.0 \\
& AF  & 0.0  & 0.0  & 0.0  & 0.0  & 0.0  & 0.0  & 0.0  & 0.0  & 0.0 \\
& CUS & 88.6 & 88.5 & 90.3 & 93.9 & 96.1 & 97.6 & 98.2 & 98.0 & 100.0 \\
\midrule
\multirow{4}{*}{LM~\cite{baumhauer2022machine}}
& R   & 89.5 & 89.3 & 90.9 & 95.1 & 96.5 & 98.1 & 98.3 & 98.0 & 100.0 \\
& NF  & 0.0  & 0.0  & 0.0  & 0.0  & 0.0  & 0.0  & 0.0  & 0.0  & 0.0 \\
& AF  & 0.0  & 0.0  & 0.0  & 0.0  & 0.0  & 0.0  & 0.0  & 0.0  & 0.0 \\
& CUS & 89.5 & 89.3 & 90.9 & 95.1 & 96.5 & 98.1 & 98.3 & 98.0 & 100.0 \\
\midrule
\multirow{4}{*}{RandRelabel~\cite{li2023random}}
& R   & 89.5 & 88.4 & 89.5 & 92.6 & 91.8 & 93.3 & 92.8 & 93.2 & 95.3 \\
& NF  & 0.0  & 0.0  & 0.0  & 0.0  & 0.0  & 0.0  & 0.0  & 0.0  & 0.0 \\
& AF  & 0.0  & 44.5 & 62.8 & 67.6 & 69.4 & 73.0 & 74.5 & 76.8 & 78.0 \\
& CUS & 89.5 & 88.4 & 89.5 & 92.6 & 91.8 & 93.3 & 92.8 & 93.2 & 95.3 \\
\midrule
\multirow{4}{*}{SSD~\cite{foster2024fast}}
& R   & 86.4 & 70.8 & 72.9 & 77.4 & 81.0 & 77.8 & 70.9 & 63.7 & 56.0 \\
& NF  & 0.0  & 42.3  & 4.2  & 37.8  & 58.3  & 94.1  & 98.2  & 85.4  & 71.4 \\
& AF  & 0.0  & 21.2 & 15.4 & 20.0 & 27.9 & 38.9 & 47.4 & 52.2 & 54.3 \\
& CUS & 86.4 & 40.8 & 69.6 & 48.1 & 33.8 & 4.59 & 1.28 & 9.3 & 16.0 \\
\midrule
\multirow{4}{*}{SalUn~\cite{fan2023salun}}
& R   & 89.5 & 88.5 & 89.3 & 92.4 & 91.9 & 93.0 & 93.1 & 93.9 & 94.8 \\
& NF  & 0.0  & 0.0  & 0.0  & 0.0  & 0.0  & 0.0  & 0.0  & 0.0  & 0.0 \\
& AF  & 0.0  & 45.7 & 62.4 & 67.7 & 69.2 & 74.0 & 74.5 & 76.8 & 78.1 \\
& CUS & 89.5 & 88.5 & 89.3 & 92.4 & 91.9 & 93.0 & 93.1 & 93.9 & 94.8 \\
\midrule
\multirow{4}{*}{\methodname{} \textbf{(Ours)}}
& R   & 89.5 & 89.1 & 90.6 & 94.9 & 96.0 & 98.0 & 98.2 & 98.2 & 100.0 \\
& NF  & 0.0  & 0.0  & 0.0  & 0.4  & 0.0  & 0.0  & 0.0  & 0.0  & 0.0 \\
& AF  & 0.0  & 0.0  & 0.0  & 0.1  & 0.0  & 0.0  & 0.0  & 0.0  & 0.0 \\
& CUS & 89.5 & 89.1 & 90.6 & 94.5 & 96.0 & 98.0 & 98.2 & 98.2 & 100.0 \\
\bottomrule
\end{tabular}
\end{threeparttable}
\end{table*}

\subsection{Qualitative Segmentation Unlearning Results}
\label{sec:supp_segqual}

\begin{figure*}[t]
    \centering
    \includegraphics[width=0.98\textwidth]{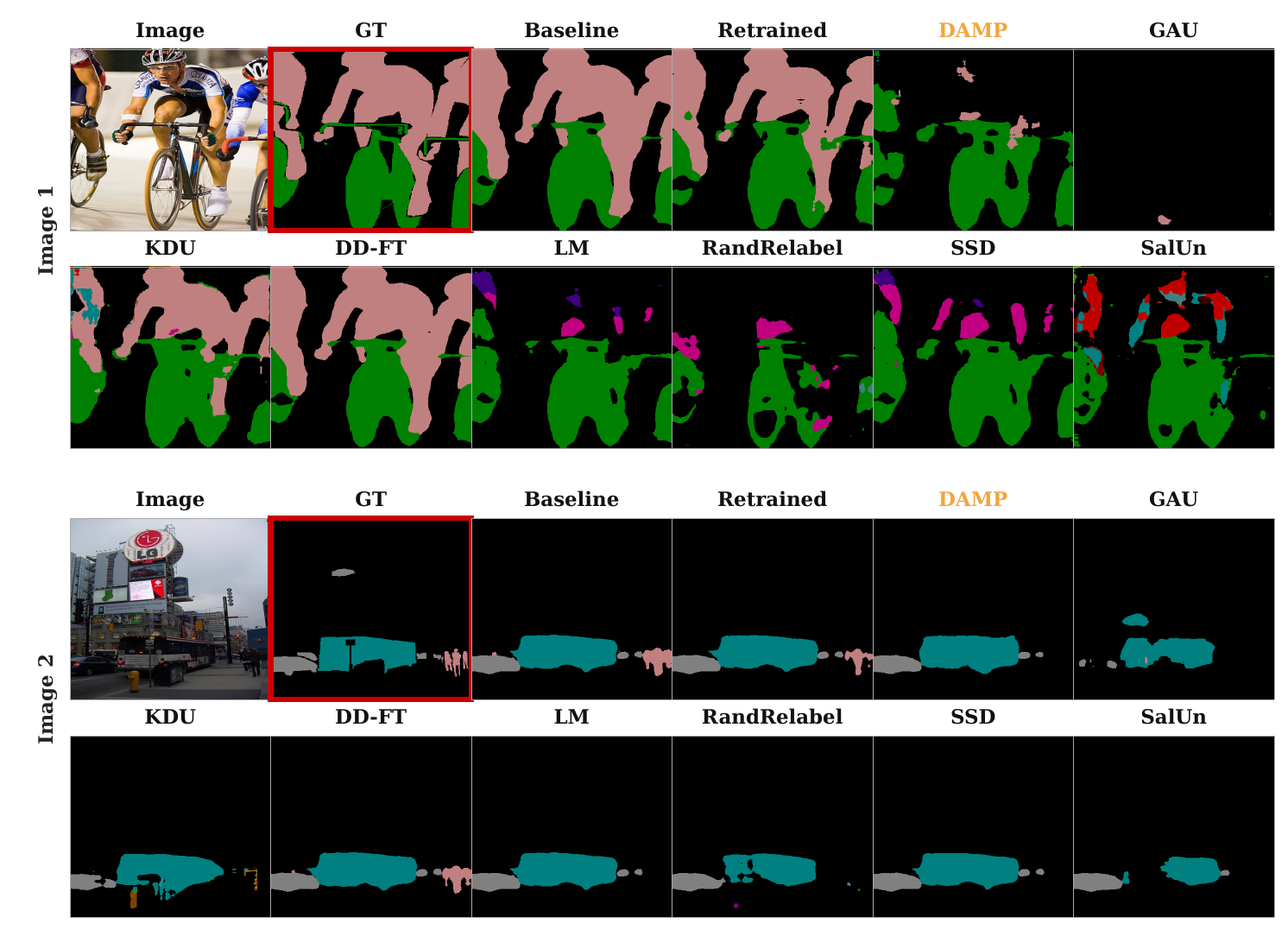}
    \caption{Qualitative results for semantic segmentation unlearning on two representative examples trying to forget Human category. Dataset is Pascal VOC 2012 semantic segmentation and the network is DeepLabV3 with a ResNet-50 backbone. We show the input image, ground-truth mask (GT), and predictions from the baseline, retraining, Gradient Ascent Unlearning (GAU), knowledge-distillation unlearning (KDU), Data Deletion Fine-Tuning (DD-FT), Logit Masking (LM), Random Relabeling~(RandRelabel), Selective Synaptic Dampening~(SSD), and Saliency Unlearning~(SalUn). \methodname{} more closely follows the retrained model by suppressing the forgotten regions while preserving the structure of retained classes, whereas several baselines either retain residual forgotten-class predictions or introduce greater degradation in the remaining segmentation regions.}
    \label{fig:seg_qual}
\end{figure*}

We further provide qualitative results for semantic segmentation unlearning in Fig.~\ref{fig:seg_qual}. The figure compares the input image, ground-truth mask, and segmentation outputs produced by the baseline model, retraining, \methodname{}, and competing baselines on representative examples. These examples highlight the visual effect of unlearning beyond aggregate accuracy metrics.

Compared with retraining, successful unlearning should remove evidence of the forgotten class while preserving the spatial structure and semantic coherence of retained regions. \methodname{} produces outputs that are visually closer to retraining, whereas several baselines either retain residual forgotten-class predictions or induce larger distortions in surrounding retained regions.

\begin{algorithm}[t]
\caption{\methodname{}: Depth-Aware Model Projection for Class Unlearning}
\label{alg:damp}
\small
\begin{algorithmic}[1]
\REQUIRE Baseline model $f_\theta$ with $L$ layers; forget set $\mathcal{F}$; retain set $\mathcal{R}$; loaders $\{\mathcal{D}_c\}_{c \in \mathcal{F}\cup\mathcal{R}}$
\ENSURE Unlearned model $f_{\theta'}$
\STATE Initialize $\theta' \leftarrow \theta$

\FOR{each class $c \in \mathcal{F}\cup\mathcal{R}$, layer $\ell$, minibatch $x \sim \mathcal{D}_c$}
    \STATE Compute edit-space vectors $z^{(\ell)}(x)$: apply GAP if $W^{(\ell+1)}$ is linear, else unfold $a^{(\ell)}(x)$ into convolutional patches matched to $W^{(\ell+1)}$
    \STATE Accumulate $s_c^{(\ell)} \mathrel{+}= \sum z^{(\ell)}(x)$, $\;n_c^{(\ell)} \mathrel{+}=$ (batch $\times$ locations)
\ENDFOR
\STATE Compute prototypes $\mu_c^{(\ell)} \leftarrow s_c^{(\ell)} / n_c^{(\ell)}$ for all $c, \ell$

\FOR{each layer $\ell = 1,\dots,L$}
    \STATE Train linear probe on GAP features $\{h_c^{(\ell)}(x)\}$ (forget vs.\ retain, 80/20 split); obtain accuracy $a_\ell$
    \STATE $\alpha_\ell \leftarrow \mathrm{clip}(2a_\ell - 1,\,0,\,1)\cdot\dfrac{\ell}{L}$
\ENDFOR

\FOR{$\ell = L, L{-}1, \dots, 1$}
    \STATE Form retain matrix $R^{(\ell)} = \big[\mu_r^{(\ell)}\big]_{r \in \mathcal{R}}$
    \STATE For each $f \in \mathcal{F}$, compute residual $d_f^{(\ell)} = \mu_f^{(\ell)} - R^{(\ell)}R^{(\ell)\dagger}\mu_f^{(\ell)}$
    \STATE Collect $Q^{(\ell)} = \{d_f^{(\ell)}/\|d_f^{(\ell)}\| : \|d_f^{(\ell)}\| > \varepsilon,\; f\in\mathcal{F}\}$
    \IF{$Q^{(\ell)} \neq \emptyset$}
        \STATE $\widetilde{Q}^{(\ell)} \leftarrow \mathrm{QR}([q_f^{(\ell)}]_{f\in\mathcal{F}})$
        \STATE $W^{(\ell+1)} \leftarrow W^{(\ell+1)}\!\left(I - \alpha_\ell\,\widetilde{Q}^{(\ell)}\widetilde{Q}^{(\ell)\top}\right)$
        \hfill\COMMENT{flatten conv weights before, reshape after}
    \ENDIF
\ENDFOR
\RETURN $f_{\theta'}$
\end{algorithmic}
\end{algorithm}